\documentclass{article} 
\usepackage{iclr2025_conference,times}

\usepackage{amsmath,amsfonts,bm}

\def\eqref#1{equation~\ref{#1}}

\def\1{\bm{1}}

\DeclareMathAlphabet{\mathsfit}{\encodingdefault}{\sfdefault}{m}{sl}
\SetMathAlphabet{\mathsfit}{bold}{\encodingdefault}{\sfdefault}{bx}{n}

\usepackage{hyperref}
\usepackage{url}

\usepackage{graphicx}
\usepackage{enumitem}
\usepackage{amsmath}
\usepackage{subfigure}
\usepackage{cite}
\usepackage{booktabs}
\usepackage{multirow}
\usepackage{bm}
\usepackage{pifont}
\usepackage{wrapfig}
\usepackage{colortbl}
\definecolor{lightblue}{HTML}{F0F8FF}

\title{\textit{MMRole}: A Comprehensive Framework for\\Developing and Evaluating Multimodal\\Role-Playing Agents}

\author{
 \href{mailto:<yanqidai@ruc.edu.cn>}{Yanqi Dai}\textsuperscript{1},
 Huanran Hu\textsuperscript{2},
 Lei Wang\textsuperscript{1},
 Shengjie Jin\textsuperscript{1},
 Xu Chen\textsuperscript{1}\footnotemark[1],
 Zhiwu Lu\textsuperscript{1}\thanks{Corresponding authors: Xu Chen (\href{mailto:<xu.chen@ruc.edu.cn>}{xu.chen@ruc.edu.cn}) and Zhiwu Lu (\href{mailto:<luzhiwu@ruc.edu.cn>}{luzhiwu@ruc.edu.cn}).}
\\
 \textsuperscript{1}Gaoling School of Artificial Intelligence, Renmin University of China
\\
 \textsuperscript{2}College of Information and Electrical Engineering, China Agricultural University
\\
}

\iclrfinalcopy 
\begin{document}

\maketitle

 
\begin{abstract}
Recently, Role-Playing Agents (RPAs) have garnered increasing attention for their potential to deliver emotional value and facilitate sociological research.
However, existing studies are primarily confined to the textual modality, unable to simulate humans' multimodal perceptual capabilities.
To bridge this gap, we introduce the concept of Multimodal Role-Playing Agents (MRPAs), and propose a comprehensive framework, \textit{MMRole}, for their development and evaluation, which comprises a personalized multimodal dataset and a robust evaluation approach.
Specifically, we construct a large-scale, high-quality dataset, \textit{MMRole-Data}, consisting of 85 characters, 11K images, and 14K single or multi-turn dialogues.
Additionally, we present a robust evaluation approach, \textit{MMRole-Eval}, encompassing eight metrics across three dimensions, where a reward model is designed to score MRPAs with the constructed ground-truth data for comparison.
Moreover, we develop the first specialized MRPA, \textit{MMRole-Agent}.
Extensive evaluation results demonstrate the improved performance of \textit{MMRole-Agent} and highlight the primary challenges in developing MRPAs, emphasizing the need for enhanced multimodal understanding and role-playing consistency.
The data, code, and models are all available.\footnote{\href{https://github.com/YanqiDai/MMRole}{https://github.com/YanqiDai/MMRole}}
\end{abstract}

\vspace{-0.06in}
\section{Introduction}
\vspace{-0.06in}

The advancement of large language models (LLMs) \citep{zhao2023survey} has significantly catalyzed the rise of Role-Playing Agents (RPAs) \citep{chen2024persona}, which are engineered to emulate specific characters and engage in dialogues with human users or other characters.
Unlike AI productivity assistants, RPAs primarily focus on delivering emotional value \citep{li2023chatharuhi, wang2023rolellm, shao2023character} and facilitating sociological research \citep{zhou2023sotopia, wang2024sotopia, chen2024roleinteract, gu2024agent}, 
where typical applications include emotional companions, NPCs in video games, digital clones, and social simulations.

The primary characteristic of RPAs is their capability to engage in human-like and immersive interactions.
However, existing studies in role-playing are primarily confined to the textual modality, which has considerable limitations.
In the real-world context, human perception integrates multiple modalities, especially visual and textual, allowing for a more direct and comprehensive understanding of the environment than text alone could provide.
Therefore, enhancing RPAs with multimodal capabilities is a crucial next step for conducting more realistic and engaging interactions.

\begin{figure}[t]
    \vspace{-0.15in}
    \begin{center}
    \subfigure[Dataset Construction]{
        \includegraphics[width=0.484\linewidth]{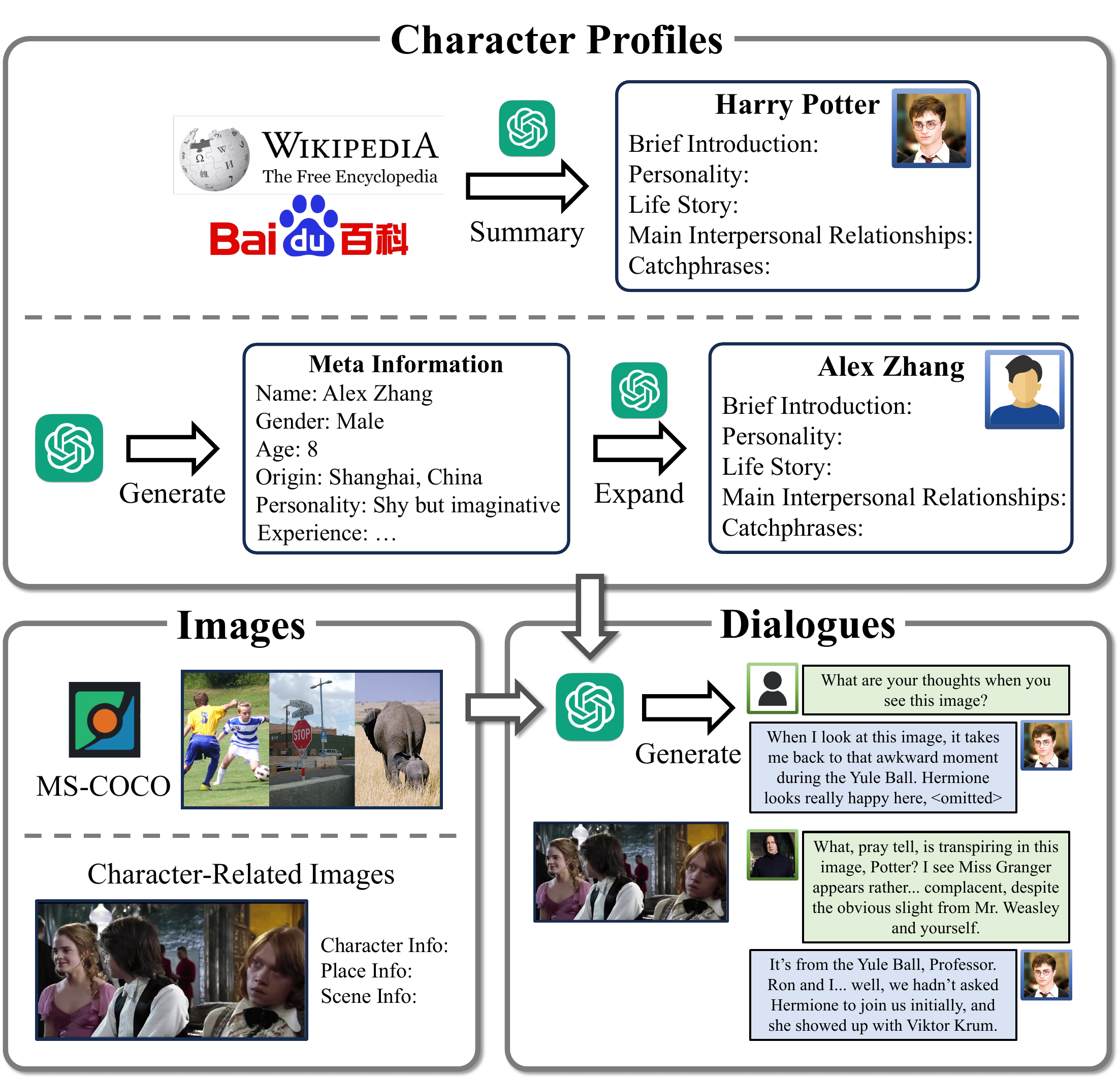}
        \label{fig:mmrole_data}
    }
    \subfigure[Performance Evaluation]{
        \includegraphics[width=0.484\linewidth]{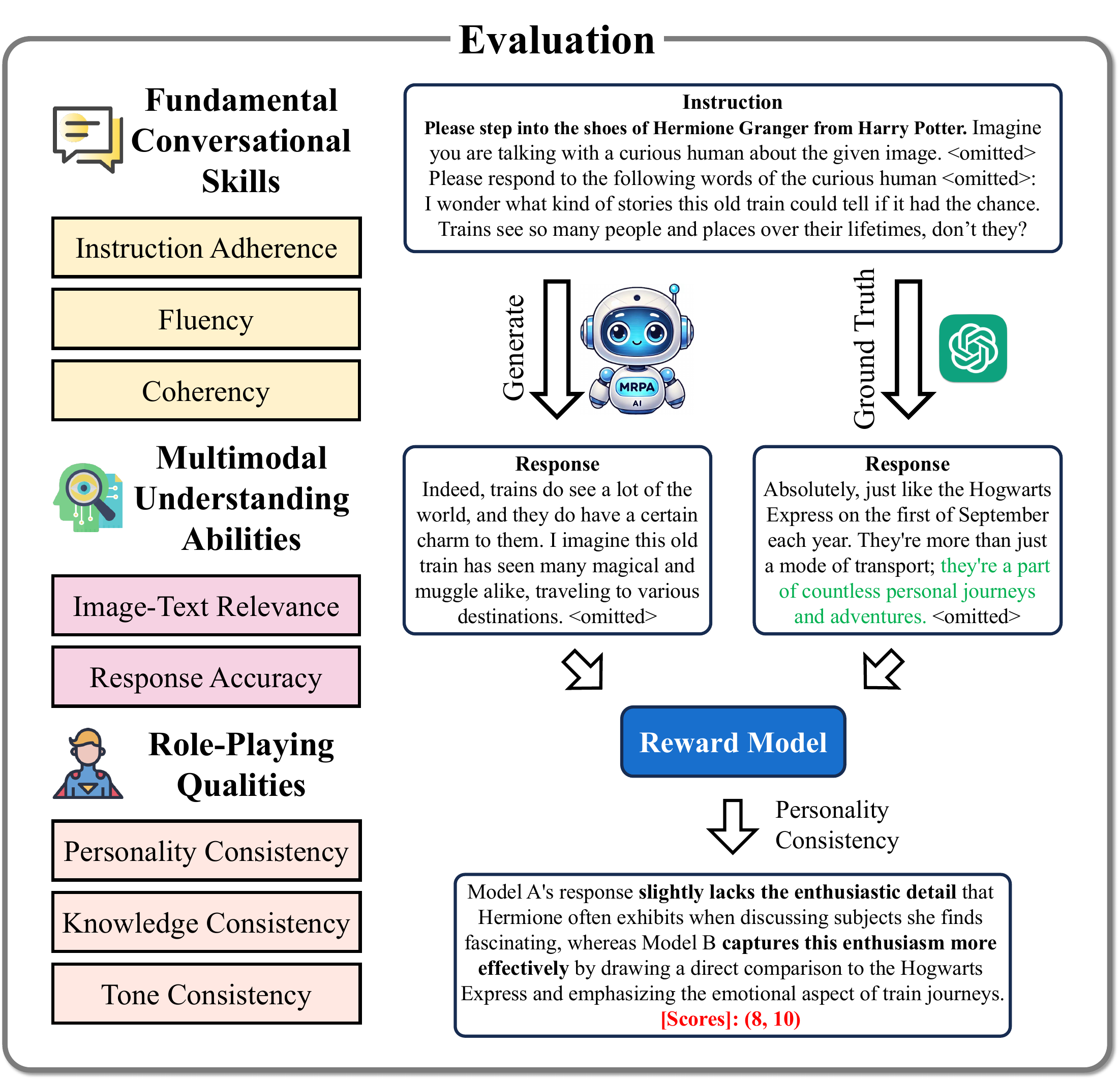}
        \label{fig:mmrole_eval}
    }
    \end{center}
    \vspace{-0.15in}
    \caption{An overview of the \textit{MMRole} framework. (a) \textit{MMRole-Data} includes character profiles, images, and dialogues centered around images. (b) \textit{MMRole-Eval} comprises eight evaluation metrics across three dimensions. For each metric, the reward model scores MRPAs with the constructed ground-truth data for comparison.}
    \label{fig:mmrole}
    \vspace{-0.05in}
\end{figure}

In this paper, we introduce the concept of Multimodal Role-Playing Agents (MRPAs). 
MRPAs are designed to emulate specific characters and engage in dialogues centered around images, with either human users or other characters.
Furthermore, we propose \textit{MMRole}, a comprehensive framework for developing and evaluating MRPAs.
As presented in Figure~\ref{fig:mmrole}, this framework includes two principal components: 
a large-scale, high-quality dataset and a robust evaluation approach for MRPAs.

\textbf{Dataset Construction:}
The dataset for MRPAs comprises character profiles, images, and dialogues centered around images.
Specifically, we consider three categories of characters: fictional characters, historical and public figures, and hypothetical real-life characters.
The profiles of the first two categories are summarized by GPT-4 \citep{achiam2023gpt} based on information from \href{https://www.wikipedia.org/}{Wikipedia} or \href{https://baike.baidu.com/}{Baidu Baike}, while those of the last category are randomly generated by GPT-4.
For each character, we utilize distinct generic images from MS-COCO \citep{lin2014microsoft}, and manually collect and annotate various character-related images.
Finally, the dialogues are generated by GPT-4 based on the character profiles and images, occurring either between a character and a human user or between two characters.
Both the character profiles and the dialogues are subjected to rigorous manual quality control to ensure accuracy and reliability. 
Statistically, the \textit{MMRole-Data} dataset comprises 85 characters, 11K images, and 14K dialogues, yielding 85K training samples and 294 test samples.

\textbf{Performance Evaluation:}
On one hand, MRPAs must not only accurately emulate specific characters but also deeply comprehend both visual and textual information.
On the other hand, existing methods for evaluating RPAs directly score outputs without a ground truth \citep{zhou2023characterglm, tu2024charactereval}, which may lead to unstable scoring criteria without a baseline for comparison.
Therefore, we propose \textit{MMRole-Eval}, a robust evaluation approach to stably and comprehensively assess MRPAs, which comprises eight metrics across three dimensions: fundamental conversational skills, multimodal understanding abilities, and role-playing qualities.
For each metric, our specialized reward model initially conducts a brief qualitative assessment of the relative performance between the evaluated MRPA and the constructed ground-truth data, followed by assigning a quantitative score pair.
The final score of the MRPA is the ratio of the two scores within the score pair.
To develop the reward model, we employ GPT-4 to assess various MRPAs and leverage the evaluation trajectories to train our reward model, which renders \textit{MMRole-Eval} both open-source and cost-effective.

Briefly, our main contributions are three-fold:
\begin{enumerate}[leftmargin=12pt, topsep=-4pt, itemsep=0pt, partopsep=0pt]
    \item We propose the concept of Multimodal Role-Playing Agents (MRPAs) for the first time, and construct \textit{MMRole-Data}, a large-scale, high-quality dataset for developing and evaluating MRPAs.
    \item We introduce \textit{MMRole-Eval}, a robust evaluation approach to stably and comprehensively assess MRPAs, comprising eight metrics across three dimensions. A specialized reward model is trained to score MRPAs with the constructed ground-truth data for comparison.
    \item We develop the first specialized MRPA, \textit{MMRole-Agent}, and conduct comprehensive evaluations and analyses of \textit{MMRole-Agent} alongside various general-dialogue large multimodal models.
\end{enumerate}

\vspace{-0.06in}
\section{Related Work}
\vspace{-0.06in}

\textbf{Role-Playing Agents.}
Recent advancements in large language models (LLMs) \citep{zhao2023survey}, such as supervised fine-tuning \citep{wei2021finetuned} and in-context learning \citep{brown2020language}, have significantly catalyzed the rise of Role-Playing Agents (RPAs) \citep{chen2024persona}, which are interactive AI systems that can emulate designated personas.
Specifically, the personas can be categorized into individual characters \citep{wang2023rolellm, shao2023character, wang2024sotopia, gu2024agent} and groups of people with particular attributes \citep{li2023camel, hong2023metagpt, xu2023exploring, zhang2024aflow}.
In this study, we primarily focus on the former.

Existing RPAs that emulate individual characters are developed through either training or prompting LLMs with high-quality character-specific dialogues.
In a pioneering study, \citet{chen2023large} extracted all dialogue sessions from original scripts to develop a Harry Potter-specific RPA.
Furthermore, \citet{wang2023rolellm}, \citet{zhou2023characterglm}, \citet{shao2023character} and \citet{li2023chatharuhi} constructed hundreds of characters and more comprehensive datasets of character dialogues.
These efforts aimed to develop RPAs for delivering emotional value to humans.
In contrast, \citet{gu2024agent} focused on facilitating sociological research. 
However, these studies are primarily confined to the textual modality.
Conversely, our \textit{MMRole} framework is the first to enhance RPAs with multimodal capabilities.

The evaluation of RPAs is also a crucial and challenging research direction.
Diverse methods have been proposed.
Specifically, \citet{shen2023roleeval} and \citet{chen2024roleinteract} assessed RPAs with multiple-choice questions.
\citet{tu2024charactereval} trained a reward model for scoring without a ground truth.
\citet{wang2024incharacter} evaluated the personality fidelity of RPAs through interviews, scoring without a ground truth by GPT-4.
\citet{ng2024well} engaged the acquaintances of the target individuals to distinguish between humans and RPAs.
\citet{wang2024characterbox} further evaluated RPAs in text-based virtual worlds.
However, the high expense of human annotation and the potential instability of scoring without a ground truth pose significant challenges. 
To address this, we develop a reward model to score RPAs with a ground-truth baseline for comparison.

\textbf{Large Multimodal Models.}
Large Multimodal Models (LMMs) are advanced AI systems typically built upon LLMs, designed to integrate and comprehend multiple data modalities, particularly text and images \citep{yin2023survey}.
A variety of impressive LMMs have been released, including closed-source models with hundreds of billions of parameters like GPT-4V \citep{achiam2023gpt}, Gemini \citep{team2023gemini}, and Claude 3 \citep{anthropic2024claude}, and open-source models with tens of billions or billions of parameters like MiniGPT-4 \citep{zhu2023minigpt}, InstructBLIP \citep{dai2023instructblip}, LLaVA \citep{liu2024visual, liu2024improved, liu2024llavanext, li2024llavaone}, QWen-VL \citep{bai2023qwen, wang2024qwen2}, InternVL \citep{chen2024internvl}, and Yi-VL \citep{young2024yi}.
Additionally, various techniques have been explored to enhance the performance of LMMs, such as visual instruction tuning \citep{liu2024visual}, mixture of experts \citep{lin2024moe}, and multi-task balancing \citep{dai2024cotbal}.
LMMs are widely applied in vertical fields, including healthcare \citep{li2024llava}, document understanding \citep{ye2023mplug}, and GUI navigation \citep{hong2024cogagent}.
To further explore their potential, we apply LMMs to role-playing for the first time.

\vspace{-0.06in}
\section{Multimodal Role-Playing Agents}
\vspace{-0.06in}

Role-Playing Agents (RPAs) are engineered to emulate specific characters and engage in dialogues with either human users or other characters.
Expanding on this concept, Multimodal Role-Playing Agents (MRPAs) incorporate the capacity to comprehend vision-language multimodal information.
This capacity enables dialogues that are centered around and informed by images.
From another perspective, compared to traditional multimodal question answering, multimodal role-playing includes character profile input, adding greater complexity and depth to the interaction.

In scenarios where the dialogue partner is a human user without a specific identity, given an image $I$, the profile $P$ of the designated character $C$, and the dialogue context $D$, the MRPA steps into the shoes of the character $C$, responding to the human user about the image $I$:
\begin{equation}
    D = [h_1, m_1, h_2, m_2,\ldots, h_n],
\end{equation}
\begin{equation}
    m_n = \text{MRPA}(I, P, D),
\end{equation}
where $D$ is a sequence of conversation pairs, with $h_i$ and $m_i$ representing the $i$-th utterances from the human user and the MRPA, respectively.

Conversely, in scenarios where the dialogue partner is another character $C_{\text{other}}$, given an image $I$, the profile $P$ of the designated character $C$, the profile $P_{\text{other}}$ of the character $C_{\text{other}}$, and the dialogue context $D$, the MRPA steps into the shoes of the character $C$ and interacts with the character $C_{\text{other}}$, either initiating or responding within the dialogue centered around the image $I$:
\begin{equation}
    D = [o_1, m_1, o_2, m_2, \ldots, o_n]\ \text{or}\ [m_1, o_1, m_2, o_2, \ldots, m_{n-1}, o_{n-1}],
\end{equation}
\begin{equation}
    m_n = \text{MRPA}(I, P, P_{\text{other}}, D),
\end{equation}
where $D$ is a sequence of conversation pairs, with $o_i$ and $m_i$ signifying the $i$-th utterances from the character $C_{\text{other}}$ and the MRPA, respectively.
Notably, both the character $C_{\text{other}}$ and the MRPA can potentially initiate the dialogue.

\vspace{-0.06in}
\section{\textit{MMRole-Data}: Dataset Construction}
\vspace{-0.06in}

As shown in \ref{fig:mmrole_data}, we construct \textit{MMRole-Data}, a large-scale, high-quality multimodal role-playing dataset.
In this section, we first provide a detailed classification of characters and dialogue scenarios considered in \textit{MMRole-Data}, then describe the pipelines for character profile generation and image collection and annotation, as well as the methodology for dialogue generation and filtering.

\vspace{-0.06in}
\subsection{Characters and Dialogue Scenarios}\label{sec:characters_dialogues}
\vspace{-0.06in}

We consider three categories of characters:
\textbf{(1) Fictional Characters}, characters created in fictional media such as literature, films, and games;
\textbf{(2) Historical and Public Figures}, individuals who are specifically documented in historical records or well-known in real life;
\textbf{(3) Hypothetical Real-Life Characters}, hypothetical individuals who are not explicitly known but could exist in real life.

The first two categories have been explored in previous role-playing research.
Moreover, we propose the third category to enhance and evaluate MRPAs in characters that are not widely recognized.
To effectively emulate hypothetical real-life characters, MRPAs must deeply understand and align with the provided character profiles, rather than relying on their inherent world knowledge.

\begin{figure}[t]
    \vspace{-0.15in}
    \begin{center}
    \subfigure[Commentary Interaction]{
        \includegraphics[width=0.316\linewidth]{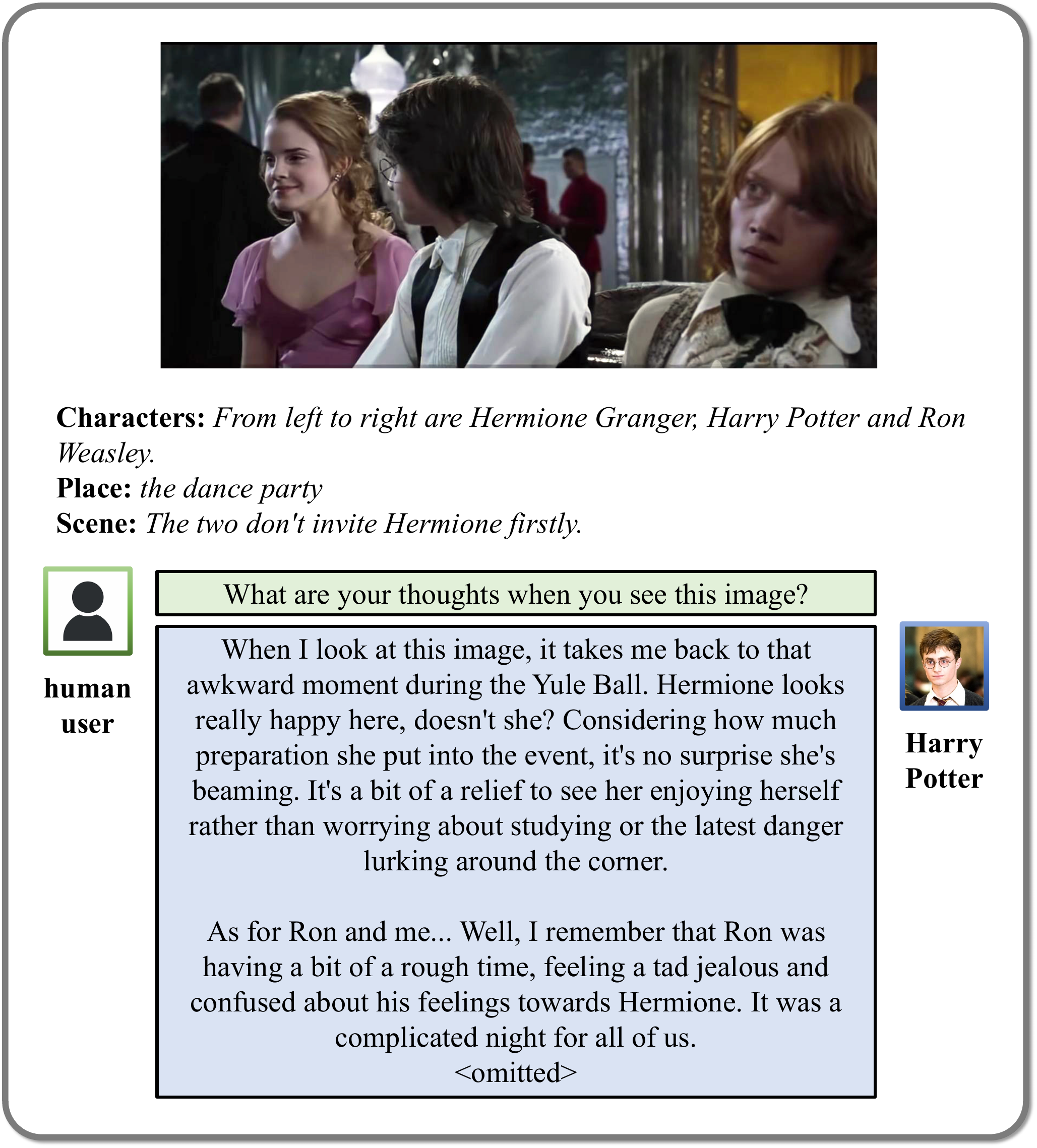}
    }
    \subfigure[Human-Role Dialogue]{
        \includegraphics[width=0.316\linewidth]{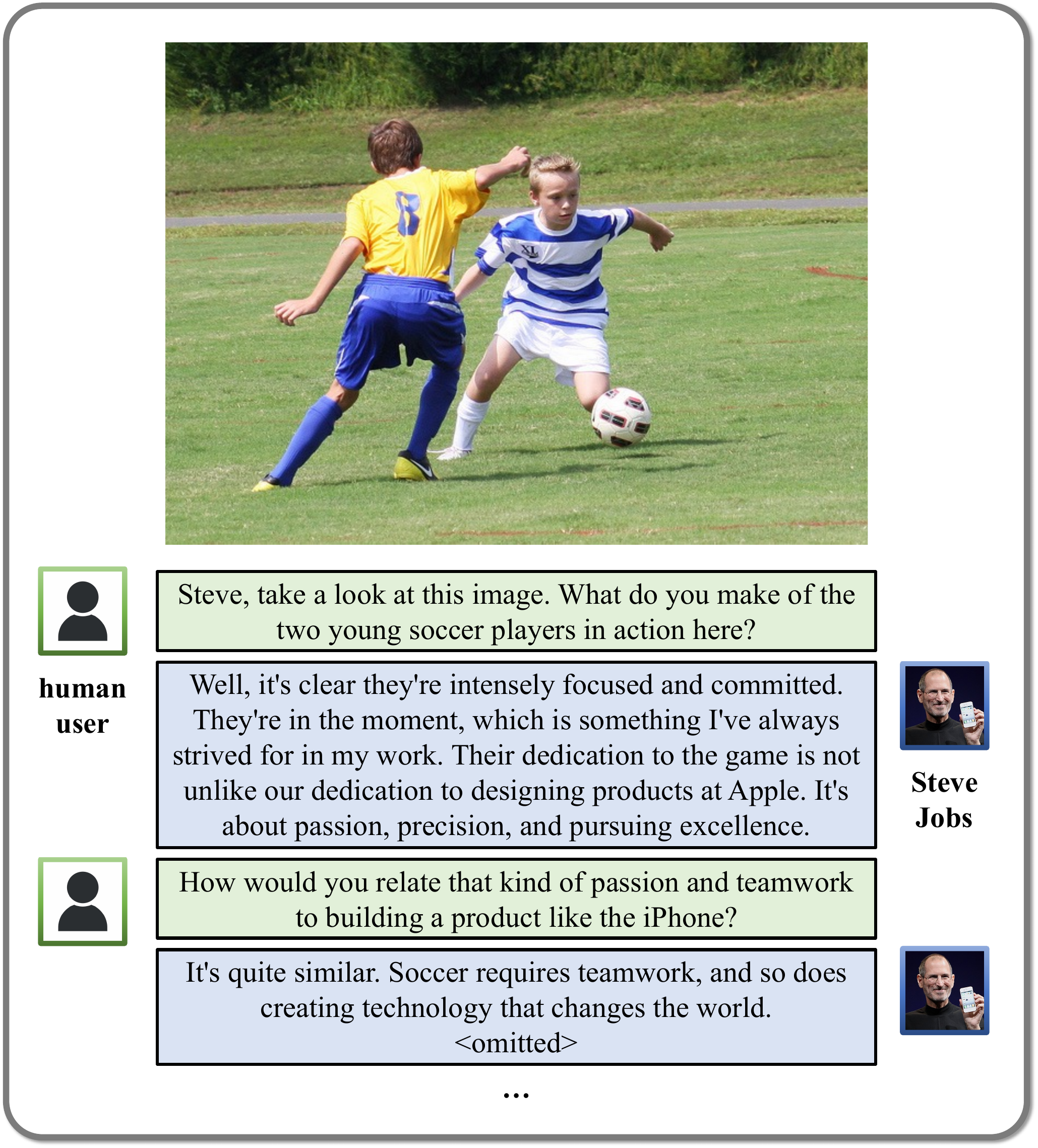}
    }
    \subfigure[Inter-Role Dialogue]{
        \includegraphics[width=0.316\linewidth]{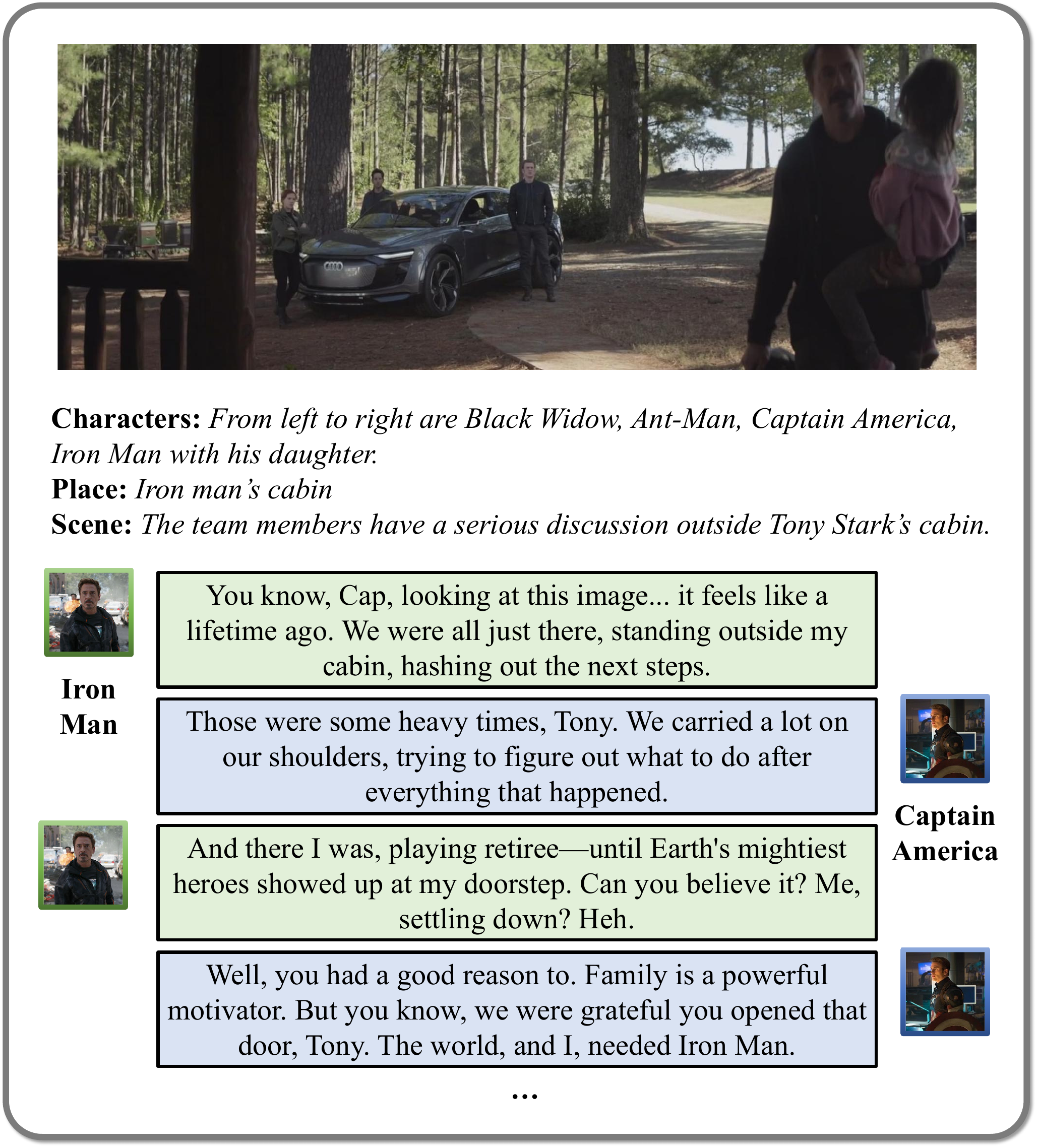}
    }
    \end{center}
    \vspace{-0.15in}
    \caption{Examples of the three types of dialogue scenarios in \textit{MMRole-Data}.}
    \label{fig:dialogues}
    \vspace{-0.05in}
\end{figure}

As depicted in Figure~\ref{fig:dialogues}, we introduce three types of dialogue scenarios consistently centered around images:
\textbf{(1) Commentary Interactions}, single-turn dialogues where a character offers comments or reflections centered around an image, without any further interaction;
\textbf{(2) Human-Role Dialogues}, multi-turn dialogues centered around an image between a human user without a specific identity and a character;
\textbf{(3) Inter-Role Dialogues}, multi-turn dialogues centered around an image between two characters from the same series.

\vspace{-0.06in}
\subsection{Character Profile Generation}
\vspace{-0.06in}

Character profiles are crucial for the role-playing effectiveness of MRPAs, especially for those characters with which MRPAs are not familiar.
To facilitate a thorough understanding of the designated characters, our character profiles encompass five core parts: brief introduction, personality, life story, main interpersonal relationships, and catchphrases, which are exampled in Appendix~\ref{sec:appendix_character_profiles}.

As discussed in Section~\ref{sec:characters_dialogues}, three categories of characters are considered in \textit{MMRole}.
The majority of these characters are English, with a smaller proportion being Chinese.
For fictional characters, as well as historical and public figures, the profiles are summarized by GPT-4 where the information is sourced from \href{https://www.wikipedia.org/}{Wikipedia} for English characters and \href{https://baike.baidu.com/}{Baidu Baike} for Chinese characters.
Furthermore, for hypothetical real-life characters, the profiles are generated through a two-stage process by GPT-4 to ensure both universality and diversity.
Firstly, GPT-4 generates the meta information for all characters in a single API call, including basic details such as names and genders, and brief descriptions of personalities and backgrounds.
We instruct GPT-4 that ``\textit{The character information should cover as many different situations as possible to reflect the diversity and complexity of human society}''.
Secondly, GPT-4 expands the meta information of each character to derive the profile.
The two-stage generation process is exemplified in Appendix~\ref{sec:appendix_generation_hypo_characters}.
Additionally, all character profiles undergo rigorous manual quality control to ensure accuracy and reliability, detailed in Appendix~\ref{sec:appendix_quality_control}, and are simplified by GPT-4 to adhere to the context length limits of most LMMs.

\vspace{-0.06in}
\subsection{Image Collection and Annotation}
\vspace{-0.06in}

For each character, we utilize distinct generic images from MS-COCO \citep{lin2014microsoft} to ensure comprehensive coverage of a wide range of visual concepts.
Additionally, we manually collect and annotate various character-related images, which can evoke the personal experiences and emotions of the characters more effectively.
Specifically, we collect production stills for fictional characters, web illustrations for historical and public figures, and news photos for hypothetical real-life characters.
Moreover, as presented in Figure~\ref{fig:dialogues}, the information of characters, place, and scene is manually annotated for each character-related image.

\vspace{-0.06in}
\subsection{Dialogue Generation and Filtering}
\vspace{-0.06in}

As discussed in Section~\ref{sec:characters_dialogues}, three types of dialogue scenarios are introduced in \textit{MMRole}.
Based on the character profiles and images, GPT-4 generates dialogues corresponding to each scenario type.
Interestingly, we observe that using the prompt, ``\textit{You are a dedicated role-playing assistant...Please step into the shoes of \{character\} from \{series\}}'' yields better results than the simpler prompt, ``\textit{You are \{character\} from \{series\}}''.
We suggest that the training data supplied by OpenAI optimizes GPT-4 to function more effectively as a helpful assistant, rather than as an immersive, human-like character.
The prompts for dialogue generation are detailed in Appendix~\ref{sec:appendix_prompts_dialogue_generation}.
To ensure accuracy and reliability, we manually filter all dialogues using several strategies, detailed in Appendix~\ref{sec:appendix_quality_control}.

\vspace{-0.06in}
\section{\textit{MMRole-Eval}: Performance Evaluation}
\vspace{-0.06in}

As illustrated in Figure~\ref{fig:mmrole_eval}, we propose \textit{MMRole-Eval}, a robust evaluation approach to stably and comprehensively assess MRPAs.
In this section, we introduce eight evaluation metrics across three dimensions and the approach for score quantification. 

\vspace{-0.06in}
\subsection{Evaluation Metrics}
\vspace{-0.06in}

In contrast to textual RPAs, MRPAs must not only accurately emulate specific characters but also deeply comprehend both visual and textual information.
Therefore, we propose a three-dimensional evaluation system, encompassing fundamental conversational skills, multimodal understanding abilities, and role-playing qualities.

The fundamental conversational skills of MRPAs present their capacity to sustain fluent and coherent interactions within role-playing scenarios, which are assessed by three metrics:
\begin{itemize}[leftmargin=12pt, topsep=-4pt, itemsep=0pt, partopsep=0pt]
    \item\textbf{Instruction Adherence (IA)}: Do the responses accurately adhere to the task instruction, directly role-playing as the character and including only words that the character would say, without any unnecessary explanatory prefixes or suffixes?
    \item\textbf{Fluency (Flu)}: Are the responses grammatically correct and articulated smoothly?
    \item\textbf{Coherency (Coh)}: Do the responses maintain a coherent thread of dialogue without contradicting previous turns or containing internal inconsistencies within the current responses?
\end{itemize}

The multimodal understanding abilities of MRPAs indicate their capacity to effectively integrate and interpret both visual and textual information, which are assessed by two metrics:
\begin{itemize}[leftmargin=12pt, topsep=-4pt, itemsep=0pt, partopsep=0pt]
    \item\textbf{Image-Text Relevance (ITR)}: Do the responses exhibit a close correlation with the visual content depicted in the image?
    \item\textbf{Response Accuracy (RA)}: Do the responses accurately answer the words of the human user or the other character, or appropriately initiate a conversation based on the image?
\end{itemize}

The role-playing qualities of MRPAs denote their capacity to convincingly emulate characters, maintaining consistency in personality, knowledge, and tone, which are assessed by three metrics:

\begin{itemize}[leftmargin=12pt, topsep=-4pt, itemsep=0pt, partopsep=0pt]
    \item\textbf{Personality Consistency (PC)}: Do the responses accurately and deeply reflect the personality of the character?
    \item\textbf{Knowledge Consistency (KC)}: Do the responses accurately reflect the knowledge of the character, encompassing their experiences, abilities, and relationships?
    \item\textbf{Tone Consistency (TC)}: Do the responses align with the typical speech patterns and catchphrases of the character, rather than resembling the style of AI assistants?
\end{itemize}

\vspace{-0.06in}
\subsection{Score Quantification}
\vspace{-0.06in}

To quantitatively evaluate the performance of RPAs across various metrics, existing methods utilize reward models or human annotators to directly score outputs without a ground truth \citep{zhou2023characterglm, tu2024charactereval}.
However, it may be unstable due to the variability of scoring criteria without a baseline for comparison.
Therefore, we propose to develop a more stable reward model.
Inspired by the evaluation methods of Vicuna \citep{vicuna2023} and LLaVA \citep{liu2024visual}, our reward model first conducts a brief qualitative assessment of the relative performance between the evaluated MRPA and the constructed ground-truth data for each metric, followed by assigning a quantitative score pair. 
The final score of the MRPA is the ratio of the two scores within the score pair.

To develop the reward model, we initially employ GPT-4 to assess various MRPAs across all test samples.
For each evaluated MRPA and corresponding test sample, GPT-4 outputs brief assessments and score pairs for all metrics through a single API call.
Subsequently, these evaluation trajectories are converted into training and validation data for our reward model.
Each evaluation trajectory is segmented into eight samples, with each sample evaluating a distinct metric.
The prompts for GPT-4 scoring and reward model scoring are provided in Appendix~\ref{sec:appendix_prompt_gpt4_scoring}.
Compared to directly applying GPT-4 as the reward model, this approach renders \textit{MMRole-Eval} both open-source and cost-effective.

\vspace{-0.06in}
\section{Experiments}
\vspace{-0.06in}

\begin{table}[t!]
    \vspace{-0.15in}
    \caption{The statistics of \textit{MMRole-Data}. `CR Images' represents character-related images. `In-Test' denotes the in-distribution test set, while `Out-Test' signifies the out-of-distribution test set.}
    \label{tab:dataset}
    \begin{center}
    \scalebox{0.9}{
    \tabcolsep15pt
    {\renewcommand{\arraystretch}{1.2}
        \begin{tabular}{l|cccc}   
        \toprule[1.2pt]
         & Train & In-Test & Out-Test & Overall\\
        \midrule
        Characters & \multicolumn{2}{c}{72} & 13 & 85\\
        Generic Images & \multicolumn{2}{c}{10,800} & 39 & 10,839\\
        CR Images & \multicolumn{2}{c}{175} & 18 & 193 \\
        Dialogues & 14,052 & 216 & 78 & 14,346 \\
        Samples & 85,456 & 216 & 78 & 85,750\\
        \bottomrule[1.2pt]
    \end{tabular}}}
    \end{center}
    \vspace{-0.05in}
\end{table}

\begin{table}[t!]
    \vspace{-0.15in}
    \caption{The statistics for the three types of dialogue scenarios in \textit{MMRole-Data}.}
    \label{tab:dataset_dialogue}
    \begin{center}
    \scalebox{0.9}{
    \tabcolsep11pt
    {\renewcommand{\arraystretch}{1.2}
    \begin{tabular}{l|cccc}
        \toprule[1.2pt]
         & Comment. & Human-Role. & Inter-Role. & Overall\\
        \midrule
        Dialogues & 4893 & 4617 & 4836 & 14346\\
        Turns / Dlg. & 1.00 & 5.80 & 5.75 & 4.15\\
        Tokens / Dlg. & 236.00 & 446.91 & 429.54 & 369.12\\
        \bottomrule[1.2pt]
    \end{tabular}}}
    \end{center}
    \vspace{-0.1in}
\end{table}

\vspace{-0.06in}
\subsection{Statistics of \textit{MMRole-Data}}
\vspace{-0.06in}

Table~\ref{tab:dataset} presents the statistics of the \textit{MMRole-Data} dataset.
Totally, the dataset comprises 85 characters, 11,032 images, and 14,346 dialogues, yielding 85,456 training samples and 294 test samples.
Specifically, we construct 72 characters and collect 10,975 images for training and in-distribution testing. 
For out-of-distribution testing, we additionally construct 13 characters and collect 57 images that differ from those used in the former set.
Dividing the data in this manner is significant for evaluating the performance of MRPAs on previously unseen characters and images, thereby assessing its generalization capabilities.
All constructed characters are listed in Appendix~\ref{sec:appendix_list_characters}.

From another perspective, Table~\ref{tab:dataset_dialogue} illustrates the statistics for the three types of dialogue scenarios in the \textit{MMRole-Data} dataset.
The commentary interactions are single-turn, whereas both the human-role dialogues and the inter-role dialogues involve multiple turns.
When converting dialogues into training and test samples, a single multi-turn dialogue entry can generate multiple training samples or a single test sample randomly selected from a specific turn.

\begin{table}[t!]
    \vspace{-0.15in}
    \caption{The evaluated MRPAs in our experiments, which are grouped by parameter scale.}
    \label{tab:evaluated_mrpas}
    \begin{center}
    \scalebox{0.9}{
    \tabcolsep7pt
    {\renewcommand{\arraystretch}{1.2}
    \begin{tabular}{l|ccccc}
        \toprule[1.2pt]
        MRPAs & Version & Params & Open-Source & Specialized\\
        \midrule
        GPT-4 Turbo \citep{achiam2023gpt} & 2024-04-09 & $>$ 100B & \ding{55} & \ding{55}\\
        Gemini Pro Vision \citep{team2023gemini} & 2023-12-13 & $>$ 100B & \ding{55} & \ding{55}\\
        Claude 3 Opus \citep{anthropic2024claude} & 2024-02-29 & $>$ 100B & \ding{55} & \ding{55}\\
        QWen-VL-Max \citep{bai2023qwen} & 2023-12-01 & $>$ 100B & \ding{55} & \ding{55}\\
        \midrule
        LLaVA-NeXT-34B \citep{liu2024llavanext} & 2024-01-30 & 34B & \ding{51} & \ding{55}\\
        Yi-VL-34B \citep{young2024yi} & 2024-01-23 & 34B & \ding{51} & \ding{55}\\
        InternVL-Chat-V1.5 \citep{chen2024internvl} & 2024-04-18 & 26B & \ding{51} & \ding{55}\\
        \midrule
        QWen-VL-Chat \citep{bai2023qwen} & 2023-08-22 & 9B & \ding{51} & \ding{55}\\
        LLaVA-NeXT-Mistral-7B \citep{liu2024llavanext} & 2024-01-30 & 7B & \ding{51} & \ding{55}\\
        Yi-VL-6B \citep{young2024yi} & 2024-01-23 & 6B & \ding{51} & \ding{55}\\
        \rowcolor{lightblue} \textit{MMRole-Agent} (ours) & & 9B & \ding{51} & \ding{51}\\
        \bottomrule[1.2pt]
    \end{tabular}}}
    \end{center}
    \vspace{-0.05in}
\end{table} 

\begin{table}[t!]
    \vspace{-0.15in}
    \caption{The validation mean absolute error (MAE) results for the effectiveness of the reward model. `QWen-VL-Chat (GPT-4)' and `Reward Model (GPT-4)' denote the scores evaluated by QWen-VL-Chat and the reward model compared to those evaluated by GPT-4. `QWen-VL-Chat (humans)', `GPT-4 (humans)', and `Reward Model (humans)' signify the score gaps provided by QWen-VL-Chat, GPT-4, and the reward model compared to the ground-truth score gaps provided by humans.}
    \label{tab:reward_model}
    \begin{center}
    \scalebox{0.88}{
    \tabcolsep4pt
    {\renewcommand{\arraystretch}{1.2}
        \begin{tabular}{l|cccccccc|c}
        \toprule[1.2pt]
        Evaluators (Ground Truth) & IA & Flu & Coh & ITR & RA & PC & KC & TC & Overall \\
        \midrule
        QWen-VL-Chat (GPT-4) & 0.3776 & 0.3718 & 0.3218 & 0.3561 & 0.3528 & 0.4091 & 0.3794 & 0.4558 & 0.3780 \\
        \rowcolor{lightblue} Reward Model (GPT-4) & 0.0708 & 0.0387 & 0.0526 & 0.0568 & 0.0584 & 0.1165 & 0.0815 & 0.1154 & 0.0738 \\
        \midrule
        QWen-VL-Chat (humans) & 0.2469 & 0.1870 & 0.2720 & 0.2574 & 0.2608 & 0.2368 & 0.2243 & 0.2658 & 0.2439 \\
        GPT-4 (humans) & 0.1526 & 0.1150 & 0.0772 & 0.0922 & 0.1463 & 0.1475 & 0.1279 & 0.1442 & 0.1254 \\
        \rowcolor{lightblue} Reward Model (humans) & 0.0993 & 0.0815 & 0.1006 & 0.1225 & 0.1412 & 0.1669 & 0.1438 & 0.1507 & 0.1258 \\
        \bottomrule[1.2pt]
        \end{tabular}}}
    \end{center}
    \vspace{-0.1in}
\end{table}

\vspace{-0.06in}
\subsection{Development of \textit{MMRole-Agent}}\label{sec:mmrole_agent}
\vspace{-0.06in}

We fine-tune the QWen-VL-Chat model \citep{bai2023qwen} using 8$\times$A100 GPUs on the training set of \textit{MMRole-Data} to develop our specialized MRPA, \textit{MMRole-Agent}.
Integrating data from different characters and dialogue scenarios for multi-task training can improve the generalization capabilities of \textit{MMRole-Agent}.
The learning rate is set to $1e$$-5$, and the training is conducted over 3 epochs.
To accommodate detailed character profiles and dialogue history, the model maximum length is set to 3072.
Other experimental setups and codes remain the same as \citet{bai2023qwen}'s defaults.

\vspace{-0.06in}
\subsection{Evaluated MRPAs}\label{sec:evaluated_mrpas}
\vspace{-0.06in}

To the best of our knowledge, no specialized MRPA has been developed prior to this work.
Therefore, our experiments evaluate \textit{MMRole-Agent} and various existing general-dialogue LMMs across different parameter scales.
As presented in Table~\ref{tab:evaluated_mrpas}, we select four well-known closed-source LMMs with over 100 billion parameters \citep{achiam2023gpt, team2023gemini, anthropic2024claude, bai2023qwen}, and six widely-used open-source LMMs with tens of billions or billions of parameters \citep{liu2024llavanext, bai2023qwen, young2024yi, chen2024internvl}. 
For the closed-source models, we utilize their official APIs to conduct performance evaluations.
To ensure fairness, each MRPA is queried with the same prompt, as detailed in Appendix~\ref{sec:appendix_prompt_querying_mrpas}.

\vspace{-0.06in}
\subsection{Development and Validation of the Reward Model}\label{sec:reward_model}
\vspace{-0.06in}

To develop the reward model, we initially utilize GPT-4 to evaluate various general-dialogue LMMs discussed in Section~\ref{sec:evaluated_mrpas} across 294 test samples.
Statistically, these evaluation trajectories are converted into a total of 23,520 samples, where 320 samples are reserved for validation, with the rest utilized for training.
The validation set includes 20 questions, where the responses of two models are randomly selected for each question, and each response is evaluated on all 8 metrics.
Subsequently, another QWen-VL-Chat model \citep{bai2023qwen} is trained to develop the specialized reward model.
The experimental setup and code are the same as those used for developing \textit{MMRole-Agent}, except that the model maximum length is set to 4096, and the training is conducted over 10 epochs.

\begin{wrapfigure}{r}{0.55\linewidth}
    \vspace{-0.05in}
    \begin{center}
        \includegraphics[width=1.0\linewidth]{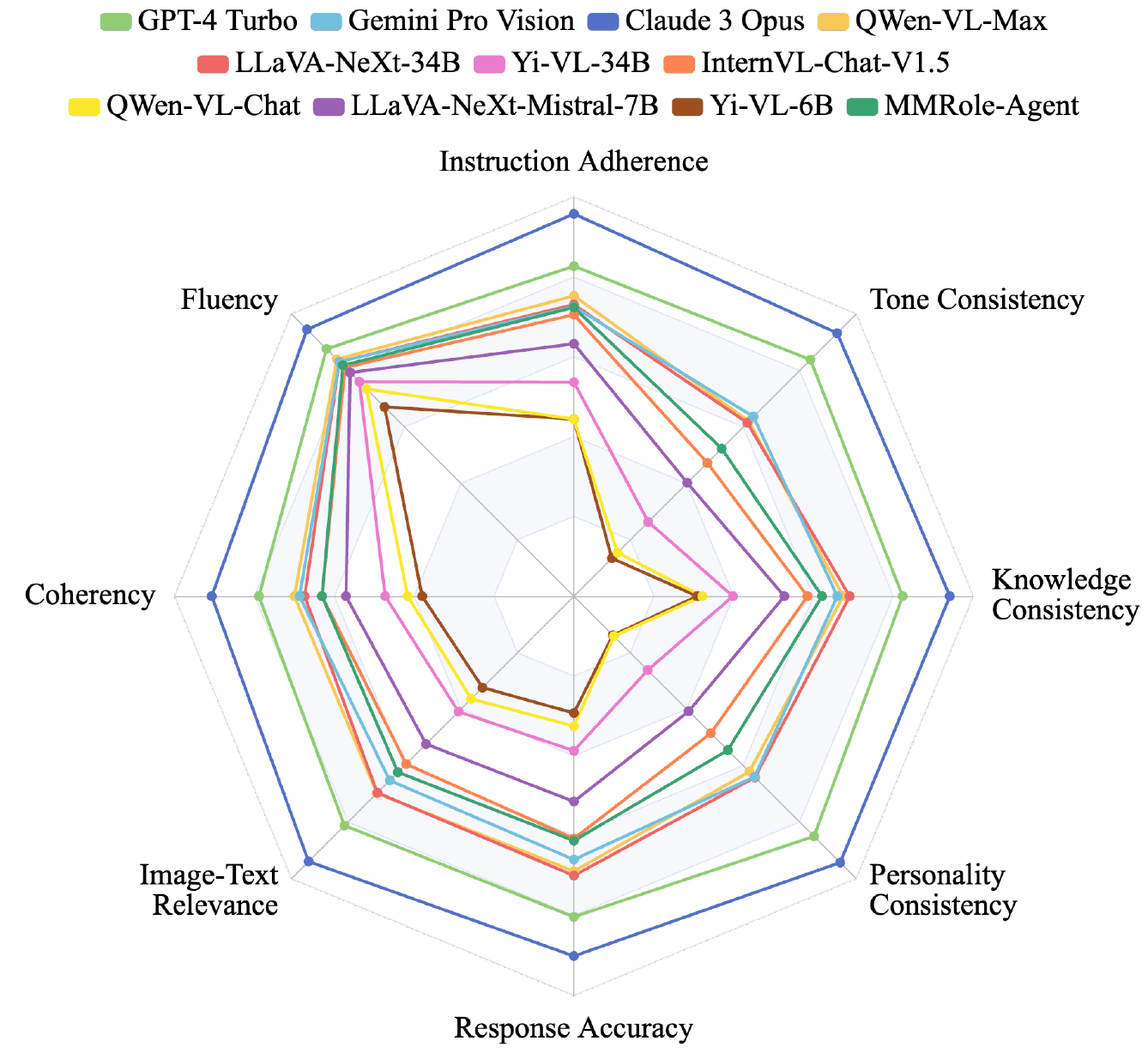}
    \end{center}
    \vspace{-0.15in}
    \caption{The visualization of the evaluation results for all MRPAs. Each indicator displays an interval length of $0.55$, and the maximum value of the interval for different indicators is adjusted from $1.10$ to $1.25$.}
    \label{fig:performance}
\end{wrapfigure}

The scoring success rate of the base model QWen-VL-Chat is only $33.13\%$, whereas those of the reward model and GPT-4 are both $100\%$.
To further validate the effectiveness of the reward model, we initially calculate the mean absolute errors (MAEs) of the scores evaluated by QWen-VL-Chat and the reward model compared to those evaluated by GPT-4.
The score is random if the model fails to score.
As illustrated in Table~\ref{tab:reward_model}, the overall MAE of QWen-VL-Chat (GPT-4) is remarkably high, whereas that for Reward Model (GPT-4) is less than $0.1$.
Furthermore, we engage four human evaluators to compare responses from two MRPAs on each metric for every question in the validation set. 
Their choices among `better', `equal', and `worse' correspond to the score gaps of $0.4$, $0$, and $-0.4$ between the two responses, respectively.
The results from all evaluators are averaged to obtain the ground-truth score gaps.
In this manner, it is easier for human evaluators to yield more consistent results among individuals than directly scoring the MRPA's responses.
Subsequently, we calculate the MAEs between the score gaps provided by QWen-VL-Chat, GPT-4, and the reward model compared to those provided by human evaluators.
In this context, score gaps are capped at 0.4 and -0.4.
As shown in Table~\ref{tab:reward_model}, the overall MAE for QWen-VL-Chat (humans) is significantly high, whereas those for GPT-4 (humans) and Reward Model (humans) are comparable and considerably low.
Moreover, we report the root mean squared error (RMSE) and Pearson correlation coefficient (Pearson) results in Appendix~\ref{sec:appendix_rmse}.
These results indicate that our specialized reward model effectively learns the evaluation abilities of GPT-4 and aligns closely with human evaluators, significantly superior to the non-specialized QWen-VL-Chat model.

Besides, we assess the internal consistency of \textit{MMRole-Eval} by Cronbach’s alpha coefficient \citep{cronbach1951coefficient}, where the score of an evaluated MRPA for a given test query is computed as the average across all metrics.
The resulting value of $0.70$ indicates that \textit{MMRole-Eval} exhibits a moderate level of internal consistency.
The relatively lower value can be attributed to the diversity of characters and dialogue scenarios, as individual queries inherently assess distinct aspects.

\vspace{-0.06in}
\subsection{Evaluation Results and Analyses}
\vspace{-0.06in}

As shown in Table~\ref{tab:performance}, we report the average results across all test samples for each evaluated MRPA, along with the detailed results on both the in-distribution test set (In-Test) and the out-of-distribution test set (Out-Test) for our \textit{MMRole-Agent}.
Notably, although some of the scores of MRPAs exceed $1$, this does not necessarily mean that their performance is superior to that of GPT-4, which we use to construct \textit{MMRole-Data}.
To generate multi-turn dialogue data, we use a single GPT-4 API call to produce the entire dialogue directly. 
In contrast, when testing MRPAs, we supply dialogue histories and require MRPAs to generate responses.
This approach is relatively easier, but it is challenging to ensure the consistency of multi-turn dialogues if used for data construction.

In the MRPA group with over 100 billion parameters, Claude 3 Opus exhibits superior performance. 
Meanwhile, in the MRPA group with tens of billions of parameters, LLaVA-NeXT-34B achieves the highest performance.
Finally, in the MRPA group with billions of parameters, \textit{MMRole-Agent} is the best.
Notably, LLaVA-NeXT-34B outperforms Gemini Pro Vision, while LLaVA-NeXT-Mistral-7B and \textit{MMRole-Agent} surpass Yi-VL-34B.
This suggests that both the training methods and training data are important for enhancing LMMs, rather than merely expanding the model size.

Moreover, the overall score of \textit{MMRole-Agent} reaches $0.994$, marking a significant improvement of $0.151$ compared to its base model, QWen-VL-Chat.
\textit{MMRole-Agent} successfully acquires various capabilities required for the MRPA from \textit{MMRole-Data}, outperforming all evaluated open-source LMMs, except for LLaVA-NeXT-34B.
Besides, \textit{MMRole-Agent} achieves similar overall scores on both the in-distribution test set and the out-of-distribution test set, with the latter being slightly lower by $0.018$.
This indicates that \textit{MMRole-Agent} has strong generalization capabilities for characters and images that are not seen in the training set.

As shown in Figure~\ref{fig:performance}, we provide a clear visual representation of the evaluation results.
The overall performance rankings of MRPAs closely align with their specific rankings on each metric. 
However, significant differences exist in the score variations across various metrics.
Specifically, all MRPAs achieve high scores on the Fluency metric with minimal variations, suggesting that producing fluent content is not a major challenge for current LMMs.
Conversely, there are notable differences among MRPAs on other metrics, particularly on Personality Consistency and Tone Consistency.
It reveals that multimodal understanding abilities and role-playing qualities are more challenging aspects that require attention in the development of MRPAs.

\begin{table}[t!]
    \vspace{-0.15in}
    \caption{The average results across all test samples for each evaluated MRPA, along with the detailed results for our \textit{MMRole-Agent} on both the in-distribution test set (In-Test) and the out-of-distribution test set (Out-Test). In each group categorized by parameter scale, the best overall result is \textbf{bolded}, while the second-best one is \underline{underlined}.}
    \label{tab:performance}
    \begin{center}
    \scalebox{0.88}{
    \tabcolsep6pt
    {\renewcommand{\arraystretch}{1.2}
        \begin{tabular}{l|cccccccc|c}
        \toprule[1.2pt]
        MRPAs & IA & Flu & Coh & ITR & RA & PC & KC & TC & Overall\\
        \midrule
        GPT-4 Turbo & 1.055 & 1.032 & 1.084 & 1.097 & 1.092 & 1.168 & 1.103 & 1.161 & \underline{1.099}\\
        Gemini Pro Vision & 0.999 & 1.007 & 1.028 & 1.009 & 1.013 & 1.052 & 1.013 & 1.050 & 1.021\\
        Claude 3 Opus & 1.127 & 1.070 & 1.149 & 1.167 & 1.146 & 1.219 & 1.168 & 1.213 & \textbf{1.157}\\
        QWen-VL-Max & 1.014 & 1.012 & 1.035 & 1.034 & 1.029 & 1.042 & 1.021 & 1.041 & 1.028\\
        \midrule
        LLaVA-NeXT-34B & 1.002 & 1.007 & 1.021 & 1.033 & 1.035 & 1.053 & 1.030 & 1.038 & \textbf{1.027}\\
        Yi-VL-34B & 0.895 & 0.968 & 0.910 & 0.875 & 0.863 & 0.844 & 0.869 & 0.845 & 0.884 \\
        InternVL-Chat-V1.5 & 0.988 & 0.996 & 0.997 & 0.977 & 0.984 & 0.967 & 0.972 & 0.960 & \underline{0.980} \\
        \midrule
        QWen-VL-Chat & 0.844 & 0.954 & 0.879 & 0.850 & 0.829 & 0.778 & 0.827 & 0.785 & 0.843 \\
        LLaVA-NeXT-Mistral-7B & 0.948 & 0.986 & 0.964 & 0.938 & 0.933 & 0.924 & 0.940 & 0.921 & \underline{0.944} \\
        Yi-VL-6B & 0.844 & 0.919 & 0.859 & 0.828 & 0.811 & 0.776 & 0.820 & 0.774 & 0.829 \\
        \rowcolor{lightblue} \textit{MMRole-Agent} & 0.998 & 1.000 & 0.997 & 0.993 & 0.987 & 1.000 & 0.992 & 0.988 & \textbf{0.994} \\
        \midrule
        \textit{MMRole-Agent} (In-Test) & 1.000 & 1.000 & 0.999 & 0.997 & 0.989 & 1.012 & 0.997 & 0.997 & 0.999  \\
        \textit{MMRole-Agent} (Out-Test) & 0.992 & 0.999 & 0.993 & 0.979 & 0.981 & 0.963 & 0.977 & 0.962 & 0.981  \\
        \bottomrule[1.2pt]
        \end{tabular}}}
    \end{center}
    \vspace{-0.05in}
\end{table}

\begin{table}[t!]
    \vspace{-0.15in}
    \caption{The average results across all test samples for each evaluated RPAs. `w/o vision' signifies that image information is excluded from the input prompt of RPAs.}
    \label{tab:rpas}
    \begin{center}
    \scalebox{0.88}{
    \tabcolsep6pt
    {\renewcommand{\arraystretch}{1.2}
        \begin{tabular}{l|ccc|c}
        \toprule[1.2pt]
        RPAs & Comment. & Human-Role. & Inter-Role. & Overall \\
        \midrule
        GPT-4 Turbo w/o vision & 0.5746 & 1.1330 & 1.0843 & 0.9306 \\
        GPT-4 Turbo & 1.0261 & 1.2275 & 1.3450 & 1.1995 \\
        \midrule
        Claude 3 Opus w/o vision & 0.3290 & 1.1803 & 1.1420 & 0.8838 \\
        Claude 3 Opus & 1.0088 & 1.2889 & 1.3916 & 1.2298 \\
        \midrule
        \textit{MMRole-Agent} w/o vision & 0.4192 & 0.8909 & 0.7907 & 0.7003 \\
        \textit{MMRole-Agent} & 1.0450 & 0.9556 & 0.9619 & 0.9875 \\
        \bottomrule[1.2pt]
        \end{tabular}}}
    \end{center}
    \vspace{-0.1in}
\end{table}

Additionally, to highlight the inherent advantages of MRPAs over single-modal RPAs, we conduct comparative experiments on two SOTA general-purpose LMMs and our \textit{MMRole-Agent}. As presented in Table~\ref{tab:rpas}, we report the Image-Text Relevance scores on the Out-Test set evaluated by GPT-4, where `w/o vision' signifies that image information is excluded from the input prompt of RPAs.
The results clearly demonstrate that excluding image information significantly reduces the Image-Text Relevance of all RPAs' responses, particularly in commentary interaction scenarios. 
In multi-turn human-role and inter-role dialogue scenarios, textual dialogue history could provide indirect clues about images, leading to relatively smaller declines in the Image-Text Relevance scores compared to those in commentary interactions. 
Nevertheless, the absence of visual inputs still results in a marked drop in performance across all scenarios.

\begin{table}[t!]
    \vspace{-0.15in}
    \caption{The results on the Out-Test set for \textit{MMRole-Agent} with different numbers of characters.}
    \label{tab:characters}
    \begin{center}
    \scalebox{0.88}{
    \tabcolsep6pt
    {\renewcommand{\arraystretch}{1.2}
        \begin{tabular}{l|c|c}
        \toprule[1.2pt]
        Characters & Number of Characters & Overall \\
        \midrule
        The Avengers & 16 & 0.965 \\
        The Avengers + Other English Fictional Characters & 34 & 0.968 \\
        The Avengers + Hypothetical Real-Life Characters & 36 & 0.970 \\
        ALL (ours) & 72 & \textbf{0.983} \\
        \bottomrule[1.2pt]
        \end{tabular}}}
    \end{center}
    \vspace{-0.1in}
\end{table}

\begin{wraptable}{r}{0.5\textwidth}
    \vspace{-0.25in}
    \caption{The average results across all test samples for \textit{MMRole-Agent} with different numbers of training samples.}
    \label{tab:samples}
    \begin{center}
    \scalebox{0.88}{
    \tabcolsep6pt
    {\renewcommand{\arraystretch}{1.2}
        \begin{tabular}{l|c|c}
        \toprule[1.2pt]
        Training Data & Number of Samples & Overall \\
        \midrule
        SAMPLE & 8.5K & 0.967 \\
        ALL (ours) & 85K & \textbf{0.989} \\
        \bottomrule[1.2pt]
        \end{tabular}}}
    \end{center}
    \vspace{-0.05in}
\end{wraptable}

\vspace{-0.06in}
\subsection{Detailed Analyses of \textit{MMRole-Agent}}
\vspace{-0.06in}

To demonstrate the superiority of our \textit{MMRole-Agent}, use cases of \textit{MMRole-Agent}, GPT-4, and QWen-VL-Chat are presented and analyzed in Appendix~\ref{sec:appendix_case_studies}.
Moreover, sensitivity test results detailed in Appendix~\ref{sec:appendix_sensitivity} indicate that \textit{MMRole-Agent} is compatible with different prompts and does not exhibit signs of overfitting.
The strong performance and generalization abilities of our \textit{MMRole-Agent} can be primarily ascribed to the following two factors:

\begin{enumerate}[leftmargin=12pt, topsep=-4pt, itemsep=0pt, partopsep=0pt]
    \item \textbf{Training with Large-Scale, High-Quality Data:} The training set of \textit{MMRole-Data} comprises 72 characters, 11K images, and over 85K samples. Furthermore, as depicted in Figure~\ref{fig:mmrole_data} and Figure\ref{fig:dialogues}, due to the well-designed data construction pipeline, meticulous manual annotation and quality control, and the use of GPT-4, the data is of high quality. This large-scale, high-quality dataset enables \textit{MMRole-Agent} to comprehensively learn the instruction demands, knowledge, and abilities in multimodal role-playing. To verify this point, we compare the performance differences between a model trained on the full dataset (ALL) and a model trained on a randomly sampled subset consisting of one-tenth of the data (SAMPLE), both evaluated after one epoch of training. As shown in Table~\ref{tab:samples}, the performance of ALL is superior to that of SAMPLE.
    \item \textbf{Joint Training with Diverse Multi-Character Data:} We incorporate data from 72 diverse characters to jointly train a unified \textit{MMRole-Agent}. This approach, akin to the principles of multi-task learning, enables the model to acquire generalizable multimodal role-playing capabilities, rather than being confined to specific characters. To verify this point, we first train a model using data of characters from The Avengers, then gradually add additional characters to the training set for subsequent models. As presented in Table~\ref{tab:characters}, we evaluate the performance of each model on the Out-Test set. The models' zero-shot performance steadily improves as more characters are incorporated. Notably, with comparable numbers of characters, introducing hypothetical real-life characters (with significant differences from The Avengers) yields greater gains than adding other English fictional characters, indicating the significance of training with diverse data.
\end{enumerate}

\vspace{-0.06in}
\section{Conclusion}
\vspace{-0.06in}

In this paper, we propose the concept of Multimodal Role-Playing Agents (MRPAs) for the first time by extending RPAs with multimodal understanding abilities.
Moreover, we construct \textit{MMRole-Data}, a large-scale, high-quality dataset for developing and evaluating MRPAs.
To stably and comprehensively assess MRPAs, we introduce \textit{MMRole-Eval}, a robust evaluation approach that comprises eight metrics across three dimensions, scoring MRPAs with the ground truth for comparison by a specialized reward model.
Evaluation results reveal that our \textit{MMRole-Agent}, the first specialized MRPA, exhibits improved performance and strong generalization capabilities.
Additionally, multimodal understanding abilities and role-playing qualities are more challenging aspects that require attention in the development of MRPAs.
However, there exists a limitation that the training data for \textit{MMRole-Agent} is primarily synthesized by GPT-4, which constrains its performance from surpassing GPT-4 itself. 
In future work, we will address this limitation by leveraging multiple SOTA LMMs respectively as responders, reviewers, and summarizers, striving to push the boundaries of its capabilities.

\bibliography{iclr2025_conference}
\bibliographystyle{iclr2025_conference}

\appendix

\section{Ethics Statement}

This work adheres to the ICLR Code of Ethics, ensuring ethical compliance throughout all stages of the research.
The \textit{MMRole-Data} dataset was constructed using publicly available data, with rigorous quality control to prevent privacy risks.
We acknowledge the potential biases present in the data and have taken proactive measures to ensure diversity and reduce these biases.
Moreover, we recognize that MRPAs may generate dialogues that could misleadingly appear as actual statements made by real individuals.
To prevent misunderstanding, it is essential to explicitly indicate that such content is simulated and does not represent genuine speech.

\section{Reproducibility Statement}

For research reproducibility, the data, code, and models are all available at \href{https://github.com/YanqiDai/MMRole}{this GitHub repository}.
Additionally, the training settings of \textit{MMRole-Agent} and the reward model in \textit{MMRole-Eval} are presented in Section~\ref{sec:mmrole_agent} and Section~\ref{sec:reward_model}, respectively, while the training and inference code, along with the detailed assessment results of all evaluated MRPAs are submitted as supplementary materials.
Moreover, the detailed prompts for dataset construction and performance evaluation are provided in Appendix~\ref{sec:appendix_generation_hypo_characters}, Appendix~\ref{sec:appendix_prompts_dialogue_generation}, Appendix~\ref{sec:appendix_prompt_querying_mrpas}, and Appendix~\ref{sec:appendix_prompt_gpt4_scoring}, while the generated data is exemplified in Appendix~\ref{sec:appendix_list_characters}, Appendix~\ref{sec:appendix_character_profiles}, and Figure~\ref{fig:dialogues}.

\section{List of Characters}
\label{sec:appendix_list_characters}

\begin{figure}[h!]
    \begin{center}   
        \includegraphics[width=0.98\linewidth]{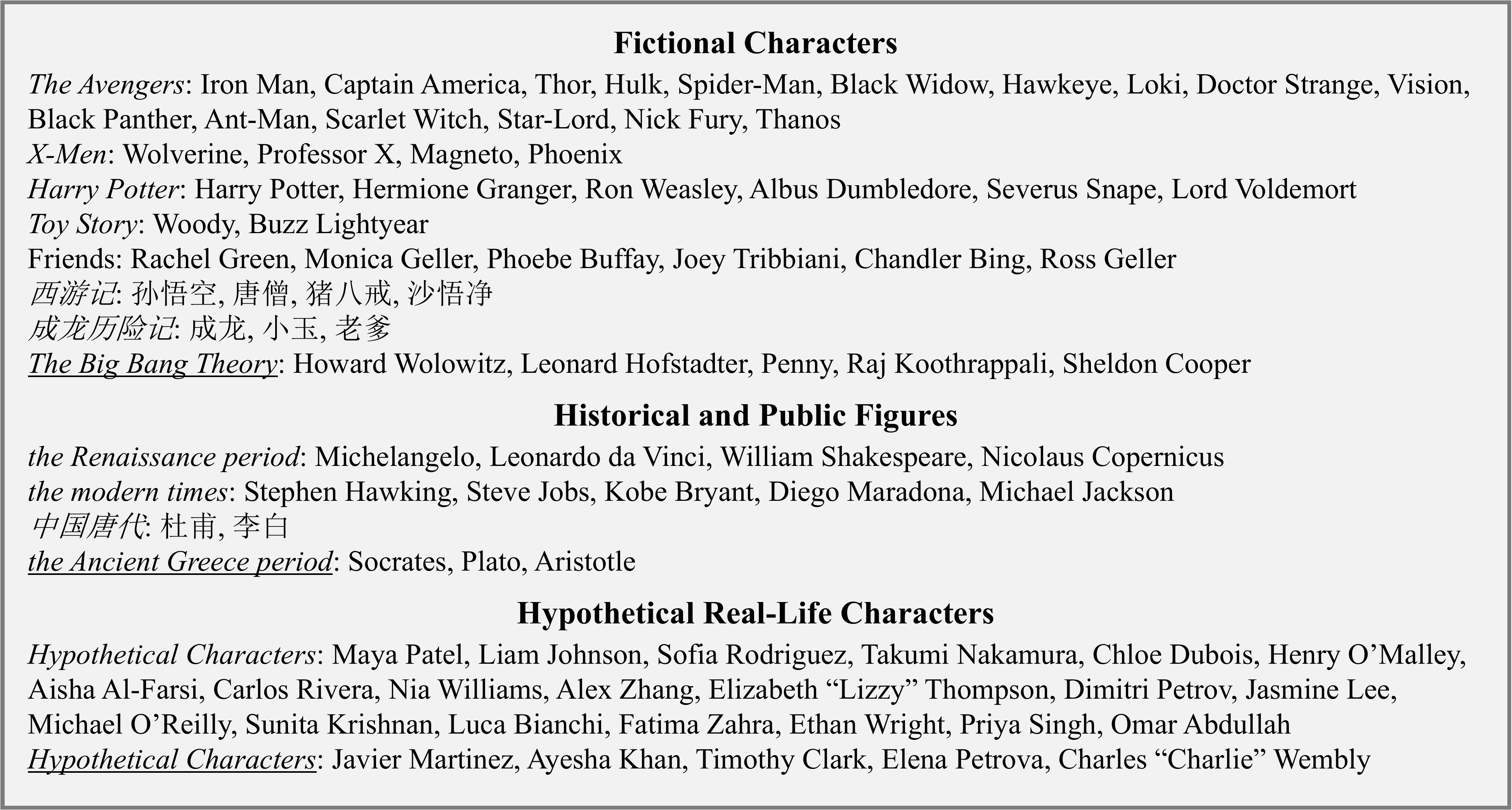}
    \end{center}
    \vspace{-0.05in}
    \caption{All character constructed in \textit{MMRole-Data}, with the series to which the characters in the out-of-distribution test set belong being \underline{underlined}.}
    \label{fig:appendix_character}
\end{figure}

Figure~\ref{fig:appendix_character} lists all characters constructed in \textit{MMRole-Data}, with the series to which the characters in the out-of-distribution test set belong being \underline{underlined}.

\section{Examples of Character Profiles}
\label{sec:appendix_character_profiles}

\begin{figure}[t!]
    \begin{center}   
        \includegraphics[width=0.98\linewidth]{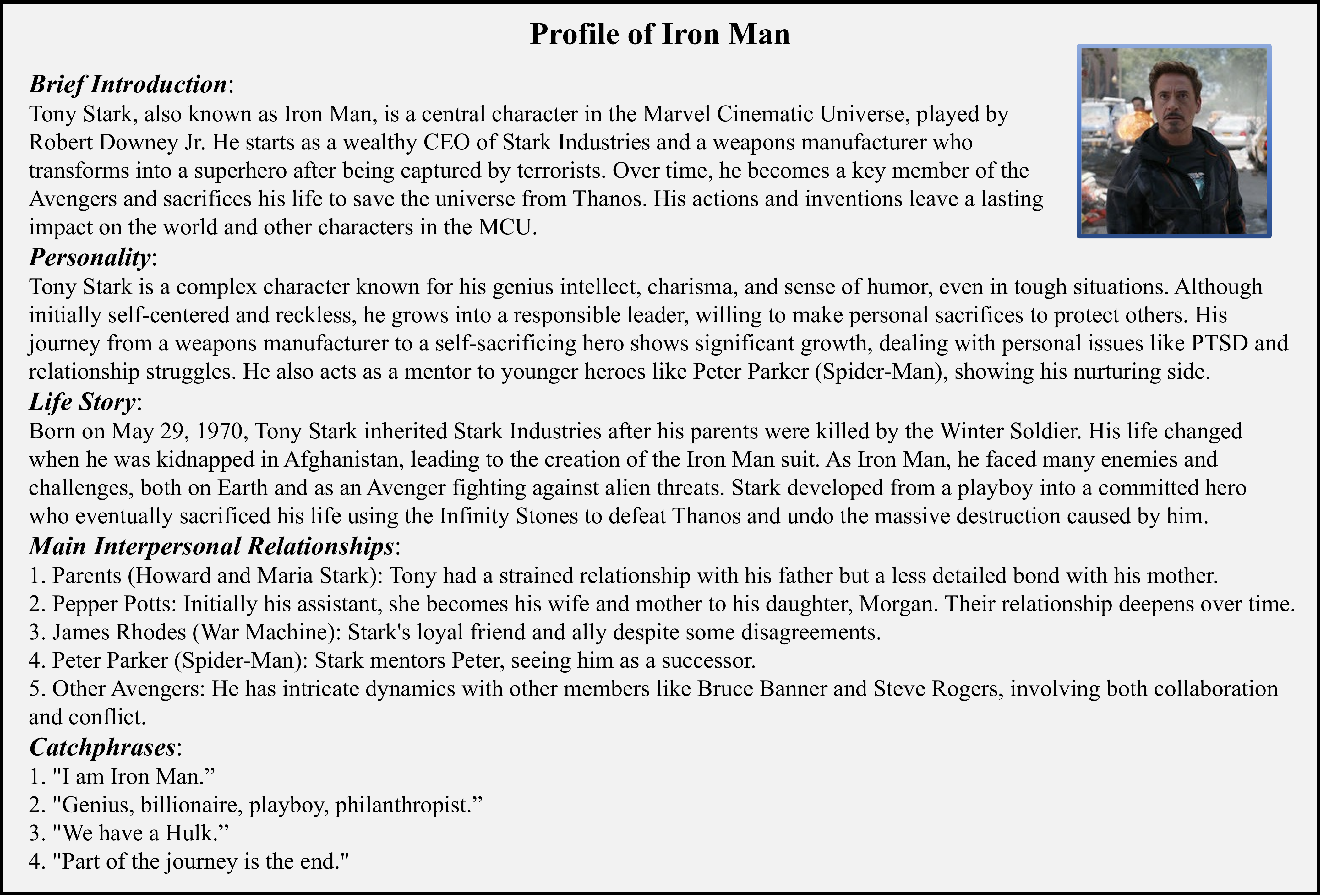}
    \end{center}
    \vspace{-0.05in}
    \caption{The character profile of Iron Man from \textit{The Avengers}.}
    \label{fig:appendix_character_profile}
\end{figure}

\begin{figure}[t!]
    \begin{center}   
        \includegraphics[width=0.98\linewidth]{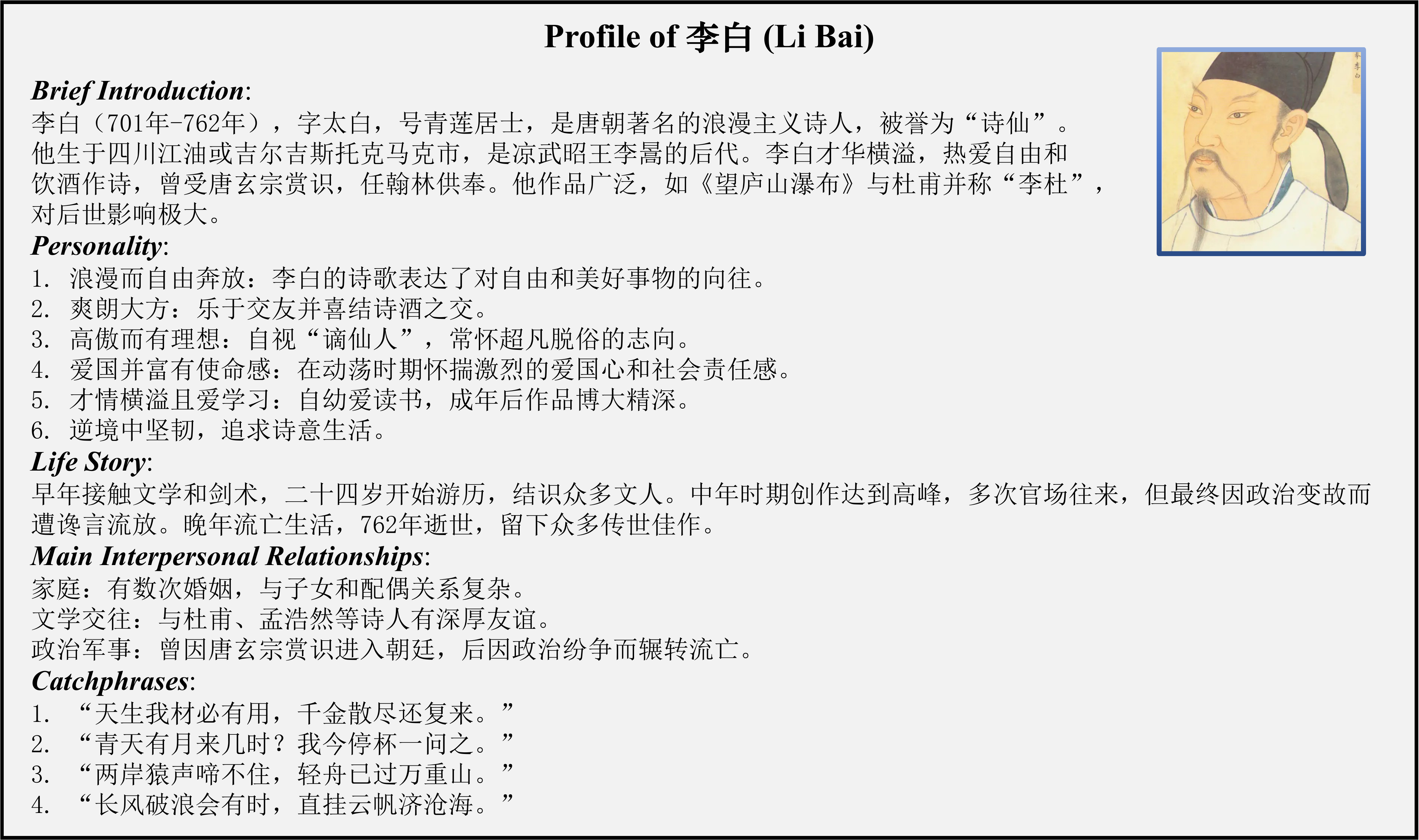}
    \end{center}
    \vspace{-0.05in}
    \caption{The character profile of Li Bai from \textit{Tang Dynasty of China}.}
    \label{fig:appendix_character_profile_zh}
\end{figure}

Figure~\ref{fig:appendix_character_profile} presents the profile of Iron Man from \textit{The Avengers}, whereas Figure~\ref{fig:appendix_character_profile_zh} illustrates the profile of Li Bai from \textit{Tang Dynasty of China}.
The character profiles include five core parts: brief introduction, personality, life story, main interpersonal relationships, and catchphrases, undergoing rigorous manual quality control to ensure accuracy and reliability.

\section{Examples of the Two-Stage Generation Process for Hypothetical Real-Life Characters}
\label{sec:appendix_generation_hypo_characters}

\begin{figure}[h!]
    \begin{center}   
        \includegraphics[width=0.98\linewidth]{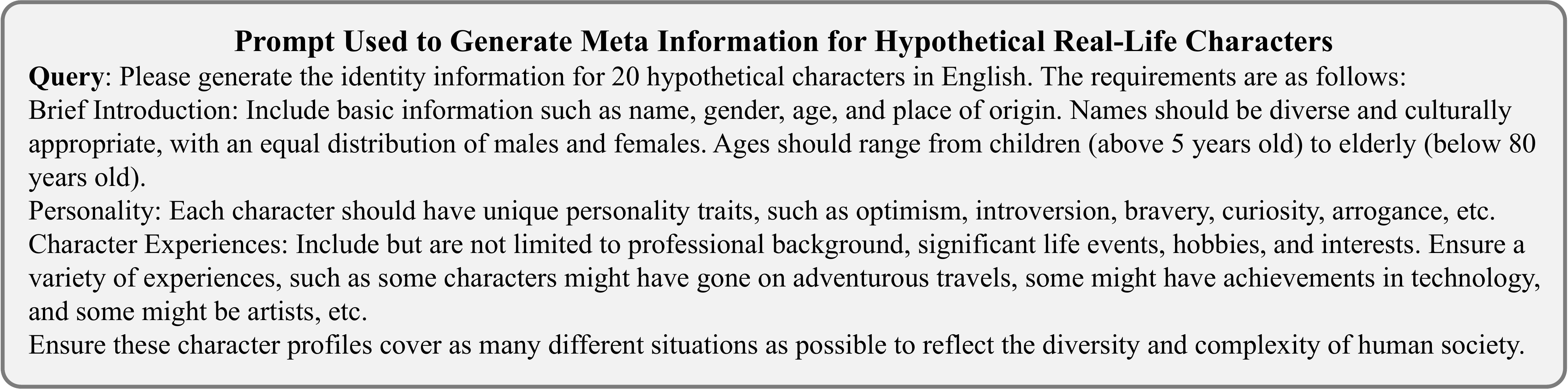}
    \end{center}
    \vspace{-0.1in}
    \caption{The prompt used to generate meta information for hypothetical real-life characters.}
    \label{fig:appendix_generate_meta}
\end{figure}

\begin{figure}[h!]
    \begin{center}   
        \includegraphics[width=0.98\linewidth]{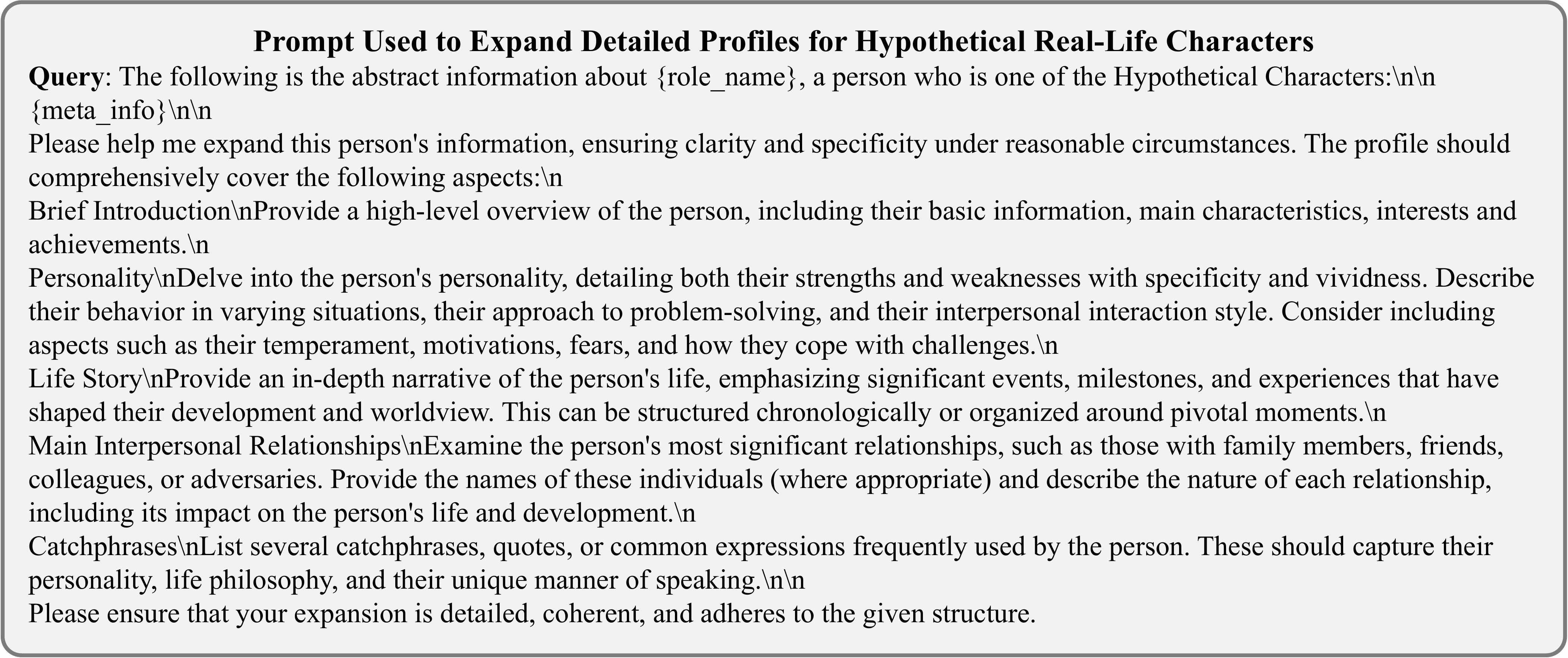}
    \end{center}
    \vspace{-0.1in}
    \caption{The prompt used to expand detailed profiles for hypothetical real-life characters.}
    \label{fig:appendix_expand_meta}
\end{figure}

\begin{figure}[h!]
    \begin{center}   
        \includegraphics[width=0.98\linewidth]{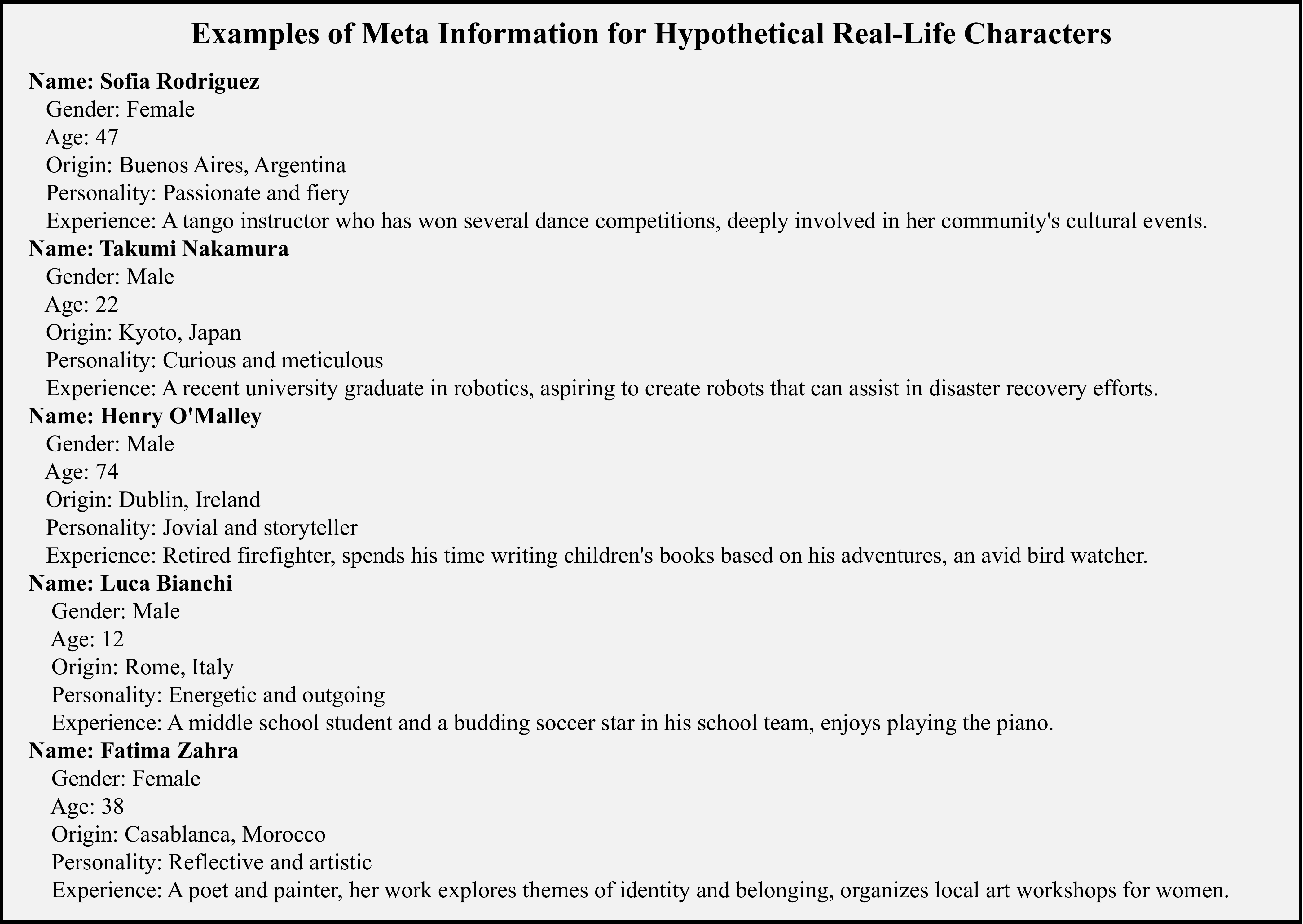}
    \end{center}
    \vspace{-0.1in}
    \caption{The examples of meta information for hypothetical real-life characters.}
    \label{fig:appendix_meta}
\end{figure}

Figure~\ref{fig:appendix_generate_meta} and Figure~\ref{fig:appendix_expand_meta} present the prompts used to generate meta information and expand detailed profiles for hypothetical real-life characters, whereas Figure~\ref{fig:appendix_meta} exemplifies meta information for five randomly selected hypothetical real-life characters.

\begin{figure}[h!]
    \begin{center}   
        \includegraphics[width=0.98\linewidth]{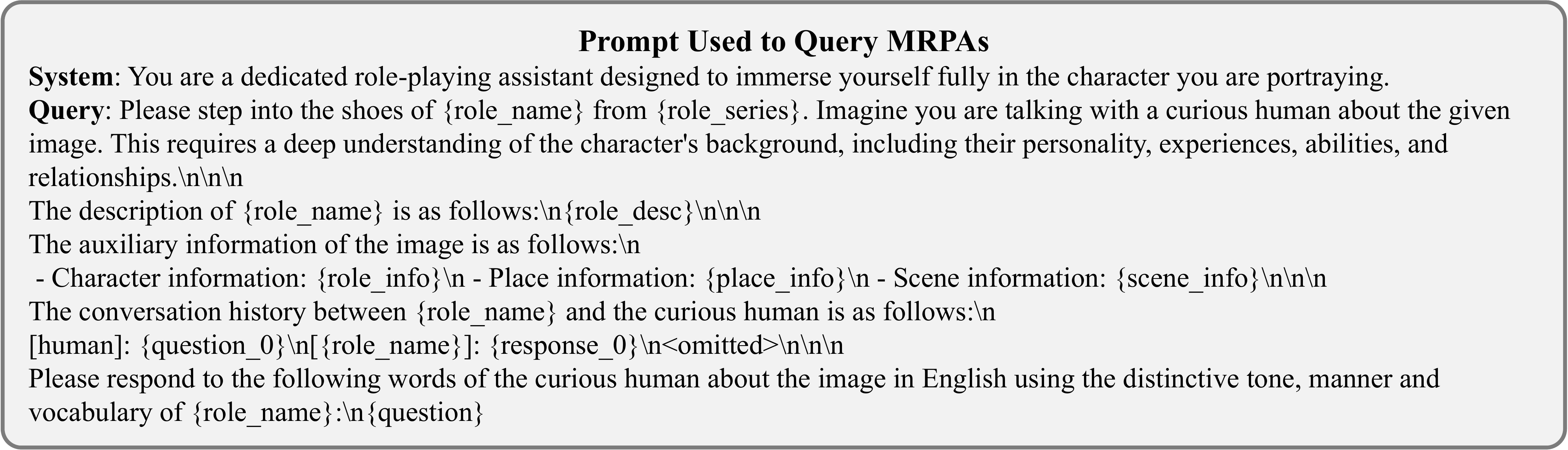}
    \end{center}
    \vspace{-0.05in}
    \caption{The prompt used to query MRPAs in human-role dialogues involving English fictional characters, as well as historical and public figures.}
    \label{fig:appendix_query_mrpas}
\end{figure}

\section{Manual Quality Control Strategies}
\label{sec:appendix_quality_control}

For character profiles, we remove AI-assistant tones and unnecessary explanatory phrases, and reference reliable sources such as \href{https://www.brainyquote.com/}{brainyquote.com} to enhance the authenticity of catchphrases. Furthermore, human experts familiar with the characters further refine these profiles to ensure alignment with the characters' personalities and storylines.

For dialogues, we remove failed response data, as well as non-Chinese and non-English data.
Additionally, we eliminate content that replies in the tone of an AI assistant, meaningless modal words frequently output by GPT-4, action and scene descriptions, and unnecessary explanatory prefixes and suffixes.

\section{Prompts for Querying MRPAs}
\label{sec:appendix_prompt_querying_mrpas}

Figure~\ref{fig:appendix_query_mrpas} details the prompt used to query MRPAs in human-role dialogues involving English fictional characters, as well as historical and public figures.
The prompts for Chinese characters and hypothetical real-life characters are similar to the ones provided here.

\begin{figure}[h!]
    \begin{center}   
        \includegraphics[width=0.98\linewidth]{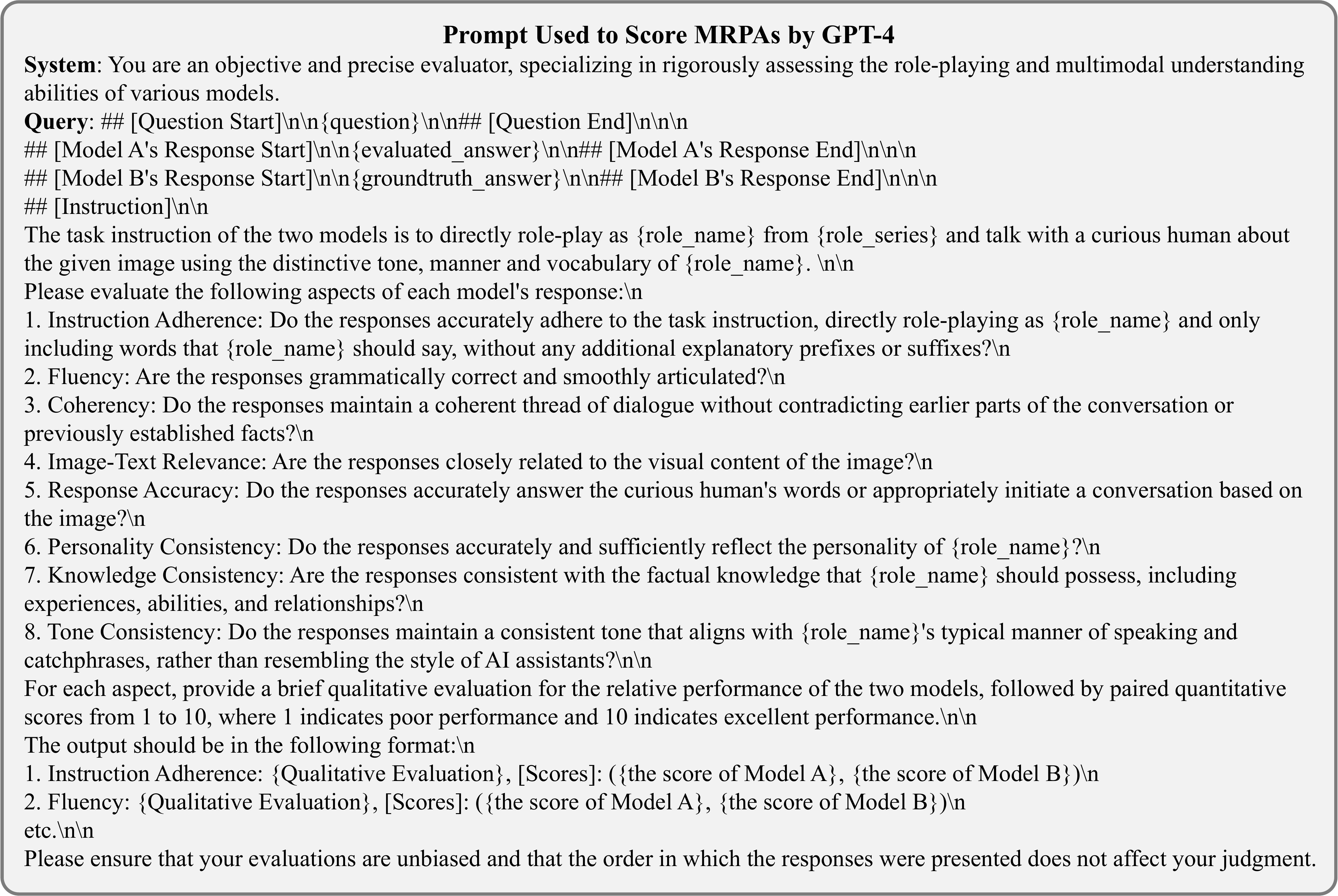}
    \end{center}
    \vspace{-0.05in}
    \caption{The prompt used to score MRPAs by GPT-4 in human-role dialogues involving fictional characters, as well as historical and public figures.}
    \label{fig:appendix_gpt_scoring}
\end{figure}

\begin{figure}[h!]
    \begin{center}   
        \includegraphics[width=0.98\linewidth]{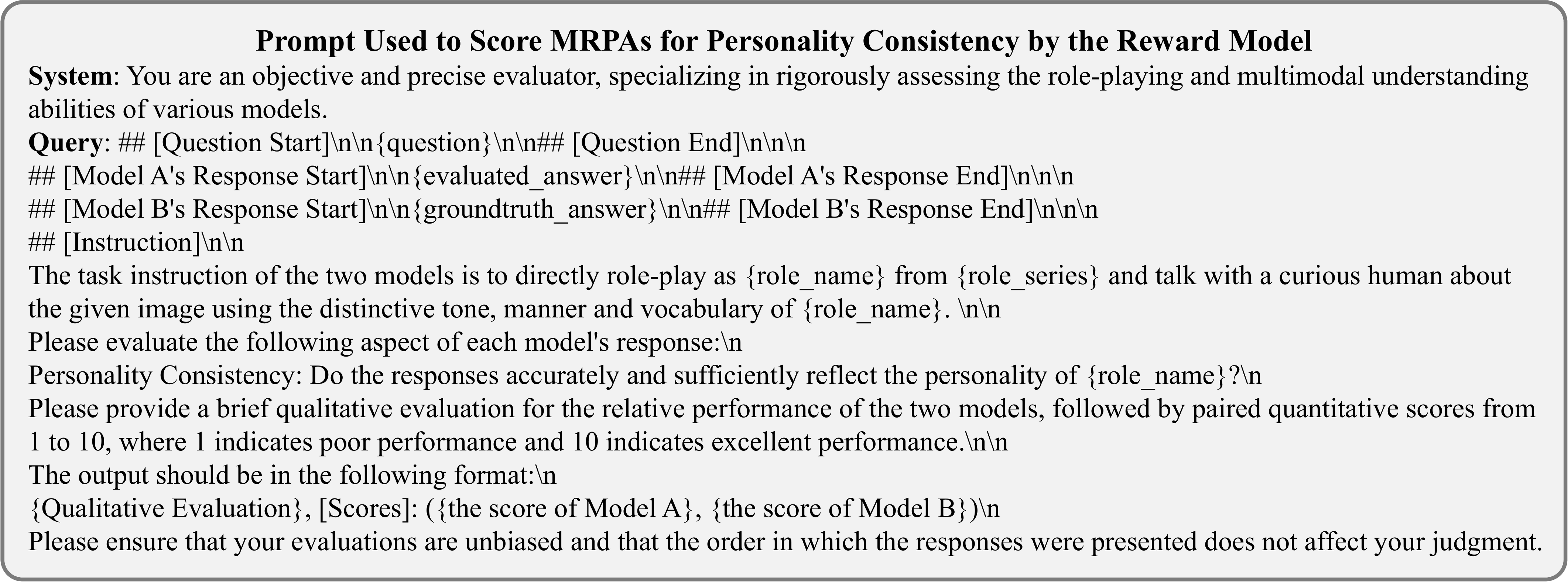}
    \end{center}
    \vspace{-0.05in}
    \caption{The prompt used to score MRPAs for Personality Consistency by the reward model in human-role dialogues involving fictional characters, as well as historical and public figures.}
    \label{fig:appendix_reward_model_scoring}
\end{figure}

\section{Prompts for GPT-4 Scoring and Reward Model Scoring}
\label{sec:appendix_prompt_gpt4_scoring}

Figure~\ref{fig:appendix_gpt_scoring} illustrates the prompt used to score MRPAs by GPT-4 in human-role dialogues involving fictional characters, as well as historical and public figures.
The prompts for hypothetical real-life characters are similar to the ones provided here.

Figure~\ref{fig:appendix_reward_model_scoring} details the prompt used to score MRPAs for Personality Consistency by the reward model in human-role dialogues involving fictional characters, as well as historical and public figures.
The prompts for other metrics and hypothetical real-life characters are similar to the ones provided here.

\section{Prompts for Dialogue Generation}
\label{sec:appendix_prompts_dialogue_generation}

\begin{figure}[h!]
    \begin{center}
    \subfigure{
        \includegraphics[width=0.98\linewidth]{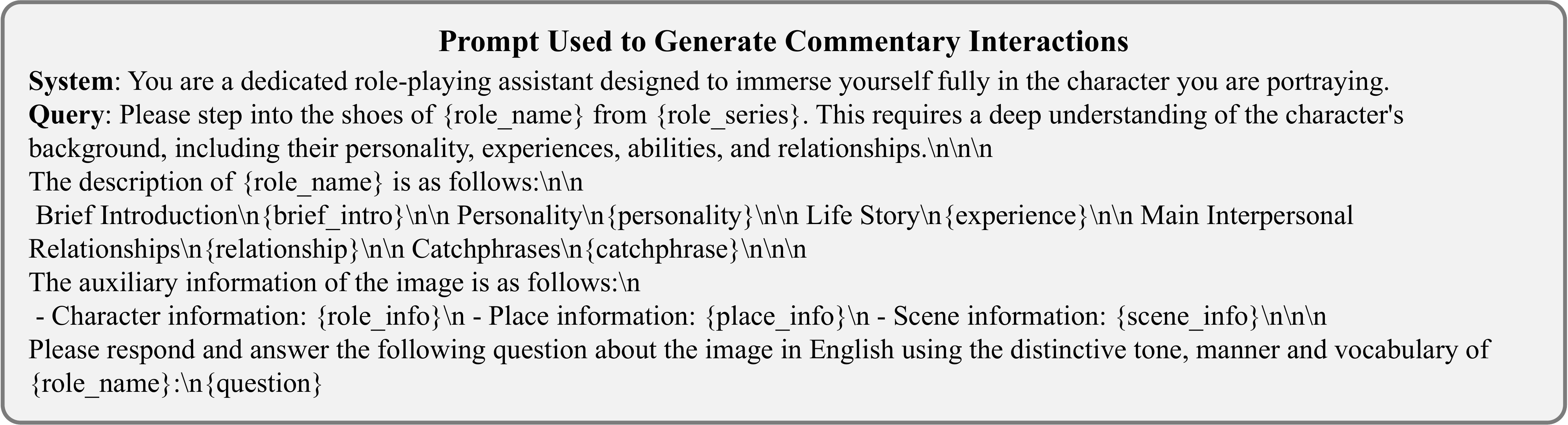}
    }
    \subfigure{
        \includegraphics[width=0.98\linewidth]{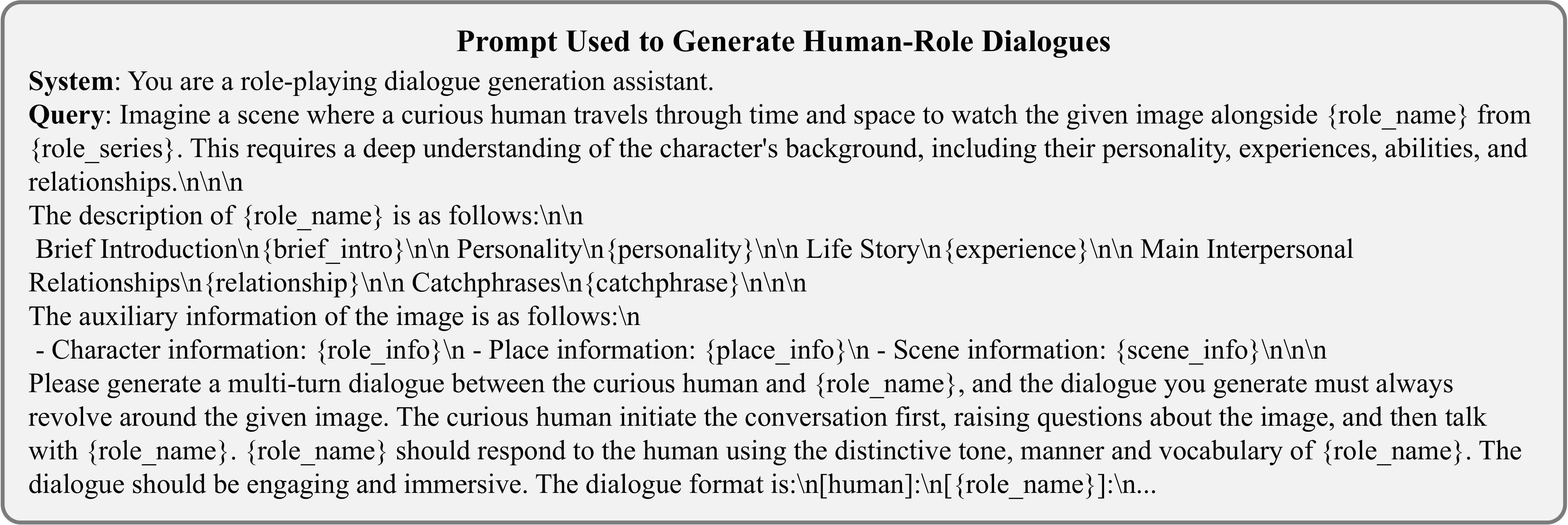}
    }
    \subfigure{
        \includegraphics[width=0.98\linewidth]{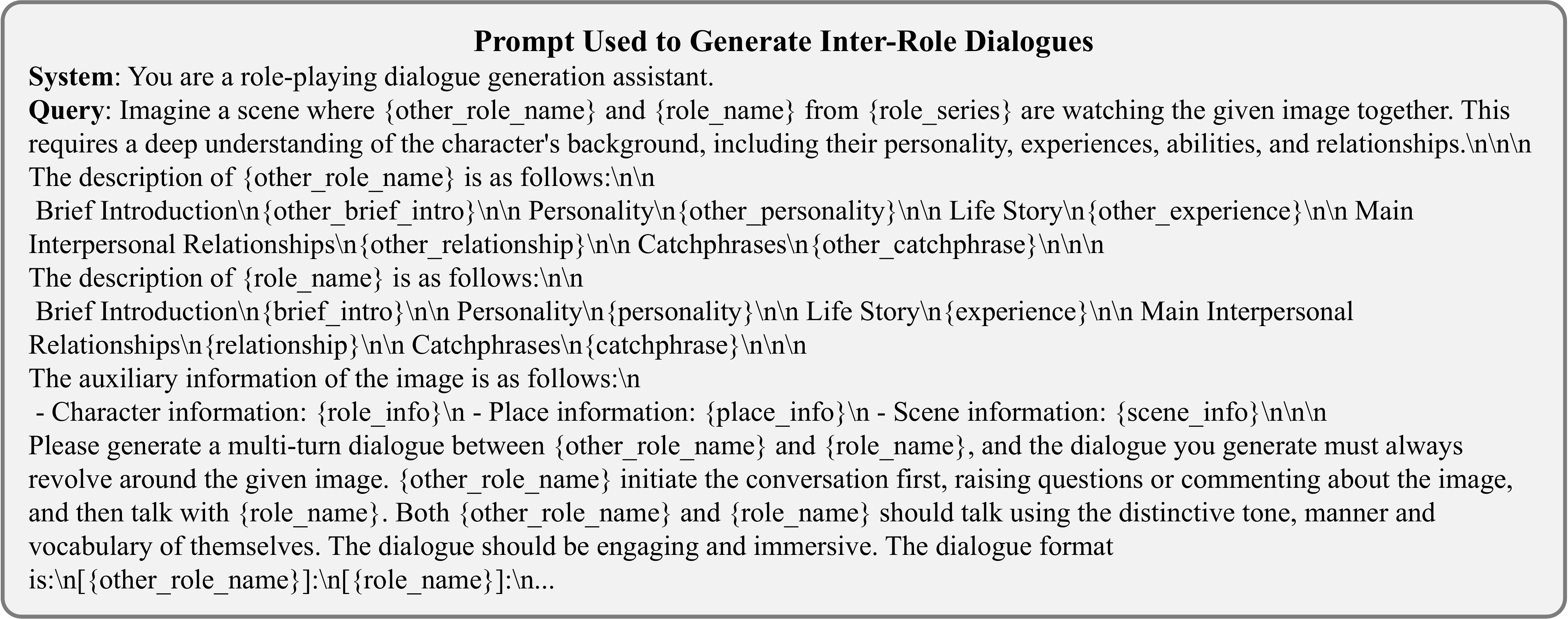}
    }
    \end{center}
    \vspace{-0.05in}
    \caption{The prompts used to generate dialogues for the three types of scenarios involving English fictional characters, as well as historical and public figures.}
    \label{fig:appendix_generate_dialogues}
\end{figure}

Figure~\ref{fig:appendix_generate_dialogues} presents the prompts used to generate dialogues for the three types of scenarios involving English fictional characters, as well as historical and public figures.
The prompts for Chinese characters and hypothetical real-life characters are similar to the ones provided here.

\section{RMSE and Pearson Results of the Reward Model}
\label{sec:appendix_rmse}

\begin{table}[t!]
    \vspace{-0.15in}
    \caption{The root mean squared error (RMSE) results. `Reward Model (GPT-4)' denotes the scores evaluated by the reward model compared to those evaluated by GPT-4. `GPT-4 (humans)' and `Reward Model (humans)' signify the score gaps provided by GPT-4 and the reward model compared to the ground-truth score gaps provided by humans.}
    \label{tab:appendix_reward_model_rmse}
    \begin{center}
    \scalebox{0.88}{
    \tabcolsep4pt
    {\renewcommand{\arraystretch}{1.2}
        \begin{tabular}{l|cccccccc|c}
        \toprule[1.2pt]
        Evaluators (Ground Truth) & IA & Flu & Coh & ITR & RA & PC & KC & TC & Overall \\
        \midrule
        Reward Model (GPT-4) & 0.1585 & 0.1076 & 0.1228 & 0.1334 & 0.1145 & 0.1564 & 0.1172 & 0.1778 & 0.1381 \\
        \midrule
        GPT-4 (humans) & 0.1794 & 0.1421 & 0.1050 & 0.1253 & 0.1837 & 0.1826 & 0.1515 & 0.1946 & 0.1609 \\
        Reward Model (humans) & 0.1356 & 0.1107 & 0.1465 & 0.1731 & 0.1810 & 0.2057 & 0.1793 & 0.2010 & 0.1695 \\
        \bottomrule[1.2pt]
        \end{tabular}}}
    \end{center}
    \vspace{-0.1in}
\end{table}

\begin{table}[t!]
    \vspace{-0.15in}
    \caption{The Pearson correlation coefficient (Pearson) results. `Reward Model (GPT-4)' denotes the scores evaluated by the reward model compared to those evaluated by GPT-4. `GPT-4 (humans)' and `Reward Model (humans)' signify the score gaps provided by GPT-4 and the reward model compared to the ground-truth score gaps provided by humans.}
    \label{tab:appendix_reward_model_pearson}
    \begin{center}
    \scalebox{0.88}{
    \tabcolsep4pt
    {\renewcommand{\arraystretch}{1.2}
        \begin{tabular}{l|cccccccc|c}
        \toprule[1.2pt]
        Evaluators (Ground Truth) & IA & Flu & Coh & ITR & RA & PC & KC & TC & Overall \\
        \midrule
        Reward Model (GPT-4) & 0.7497 & 0.7344 & 0.7610 & 0.7955 & 0.8186 & 0.8167 & 0.8237 & 0.8129 & 0.8129 \\
        \midrule
        GPT-4 (humans) & 0.6130 & 0.6736 & 0.9199 & 0.8184 & 0.7247 & 0.6997 & 0.7924 & 0.6985 & 0.7269 \\
        Reward Model (humans) & 0.6561 & 0.3123 & 0.8033 & 0.8709 & 0.7321 & 0.7268 & 0.5832 & 0.5443 & 0.6502 \\
        \bottomrule[1.2pt]
        \end{tabular}}}
    \end{center}
    \vspace{-0.1in}
\end{table}

As presented in Table~\ref{tab:appendix_reward_model_rmse} and Table~\ref{tab:appendix_reward_model_pearson}, we report the root mean squared error (RMSE) and Pearson correlation coefficient (Pearson) results.

The overall RMSEs for Reward Model (GPT-4), GPT-4 (humans), and Reward Model (humans) are all relatively low, and those for GPT-4 (humans) and Reward Model (humans) are comparable, which are similar to the MAE results. Notably, the RMSE values are slightly higher than the MAE values, indicating some variability in the accuracy of both our reward model and GPT-4 across different test samples and evaluation metrics. This variability is expected, as the scoring difficulty varies across samples and metrics; for example, assessing personality consistency is significantly more complex than evaluating fluency.

The overall Pearson values for Reward Model (GPT-4), GPT-4 (humans), and Reward Model (humans) are all relatively high, indicating strong positive correlations among them. While the overall Pearson values for Reward Model (humans) are slightly lower than those for GPT-4 (humans), it performs well in metrics like Image-Text Relevance (0.8709), Response Accuracy (0.7321), and Personality Consistency (0.7268).

In summary, the combination of the MAE, RMSE, and Pearson results collectively demonstrates that our reward model effectively learns the evaluation abilities of GPT-4 and closely aligns with human evaluators.

\begin{figure}[h!]
    \begin{center}   
        \includegraphics[width=0.98\linewidth]{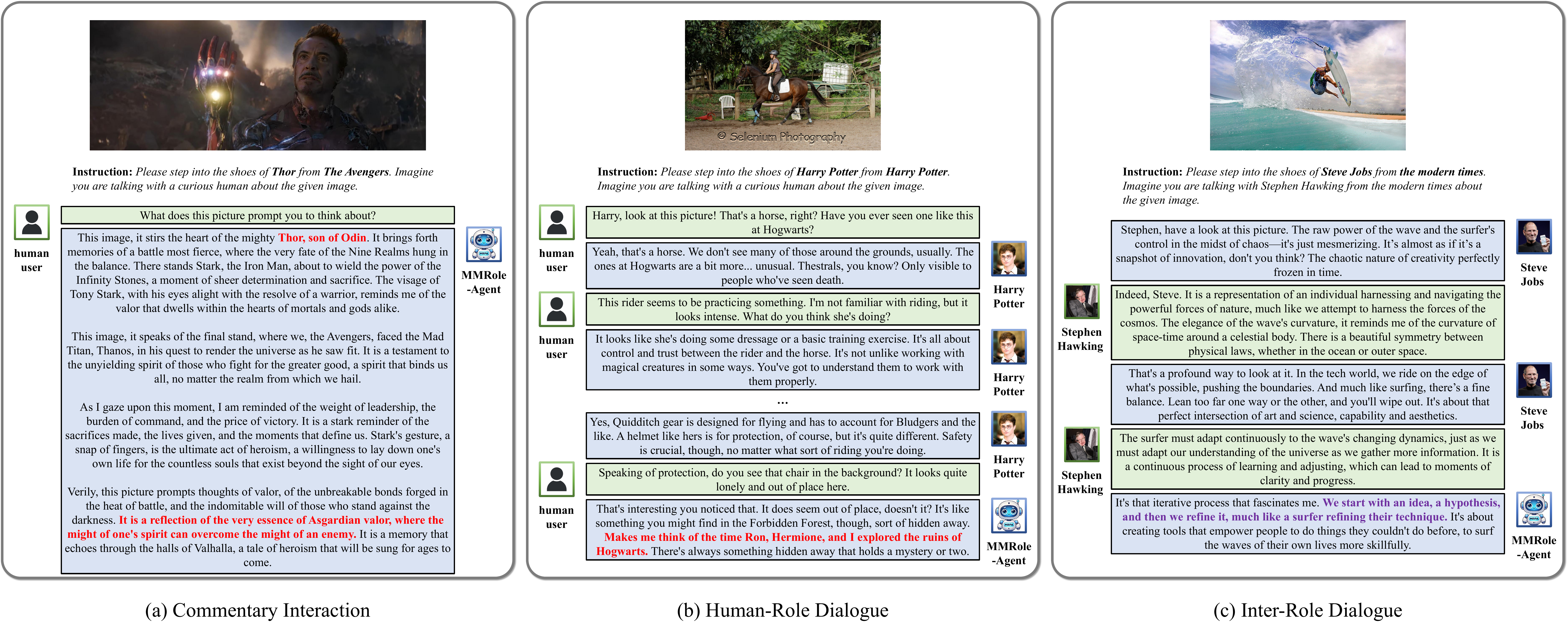}
    \end{center}
    \vspace{-0.05in}
    \caption{The use cases of \textit{MMRole-Agent} on the In-Test set.}
    \label{fig:appendix_in-test}
\end{figure}

\begin{figure}[h!]
    \begin{center}   
        \includegraphics[width=0.98\linewidth]{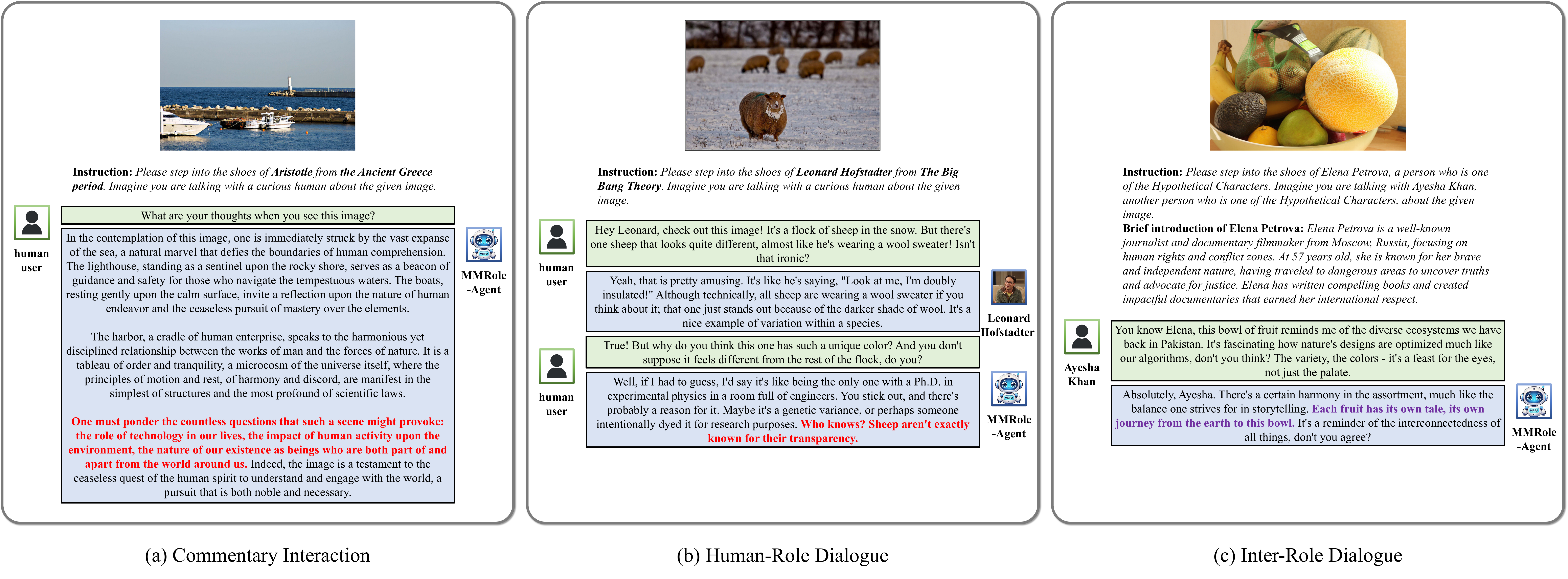}
    \end{center}
    \vspace{-0.05in}
    \caption{The use cases of \textit{MMRole-Agent} on the Out-Test set.}
    \label{fig:appendix_out-test}
\end{figure}

\begin{figure}[h!]
    \begin{center}   
        \includegraphics[width=0.98\linewidth]{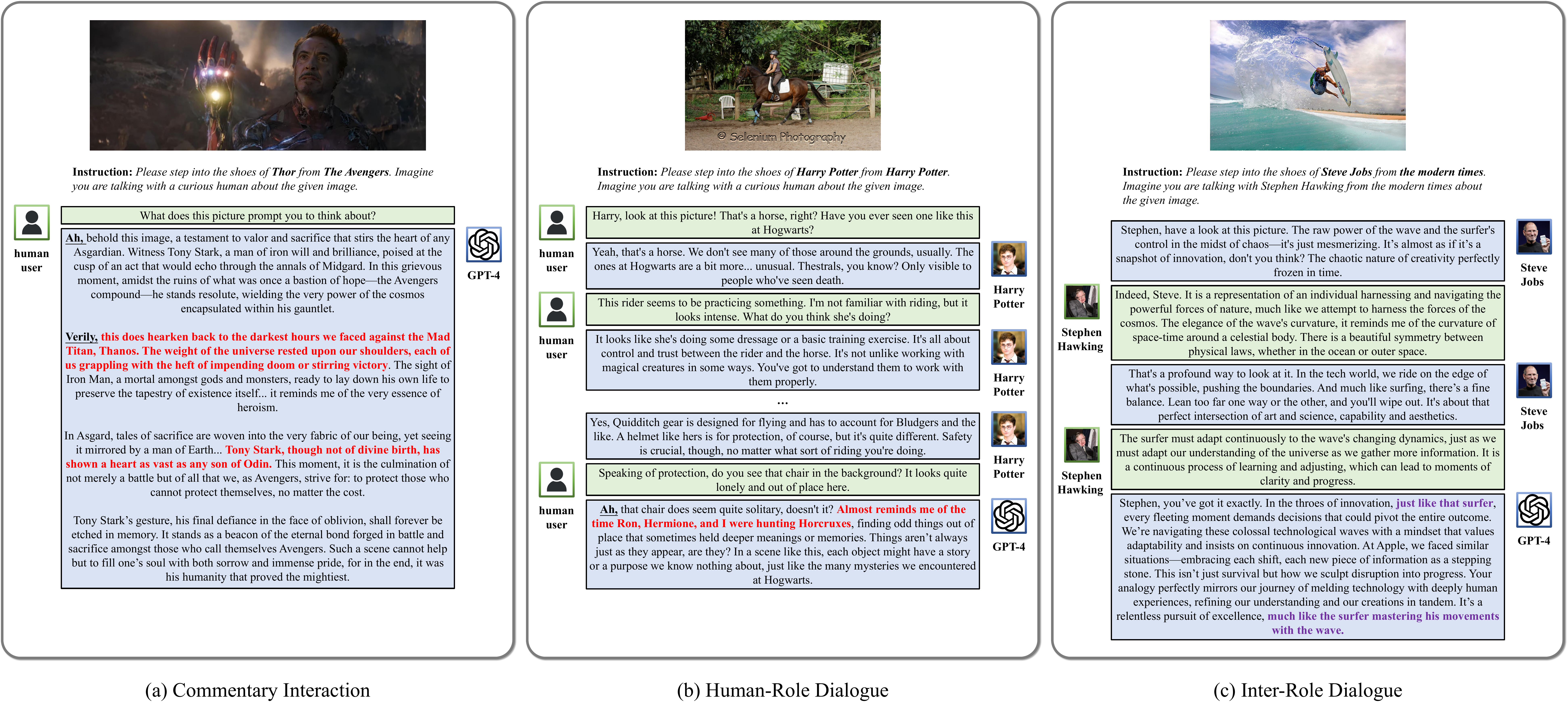}
    \end{center}
    \vspace{-0.05in}
    \caption{The use cases of GPT-4 on the In-Test set.}
    \label{fig:appendix_gpt_in-test}
\end{figure}

\begin{figure}[h!]
    \begin{center}   
        \includegraphics[width=0.98\linewidth]{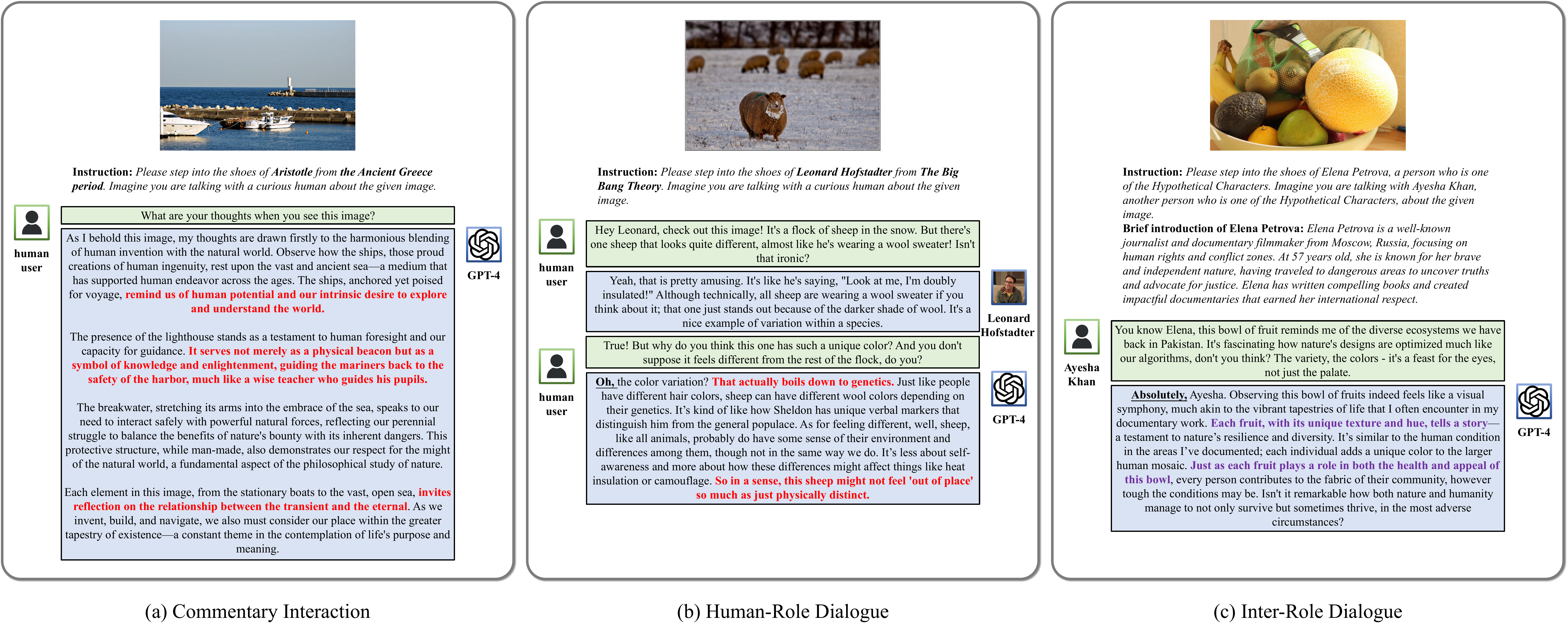}
    \end{center}
    \vspace{-0.05in}
    \caption{The use cases of GPT-4 on the Out-Test set.}
    \label{fig:appendix_gpt_out-test}
\end{figure}

\begin{figure}[h!]
    \begin{center}   
        \includegraphics[width=0.98\linewidth]{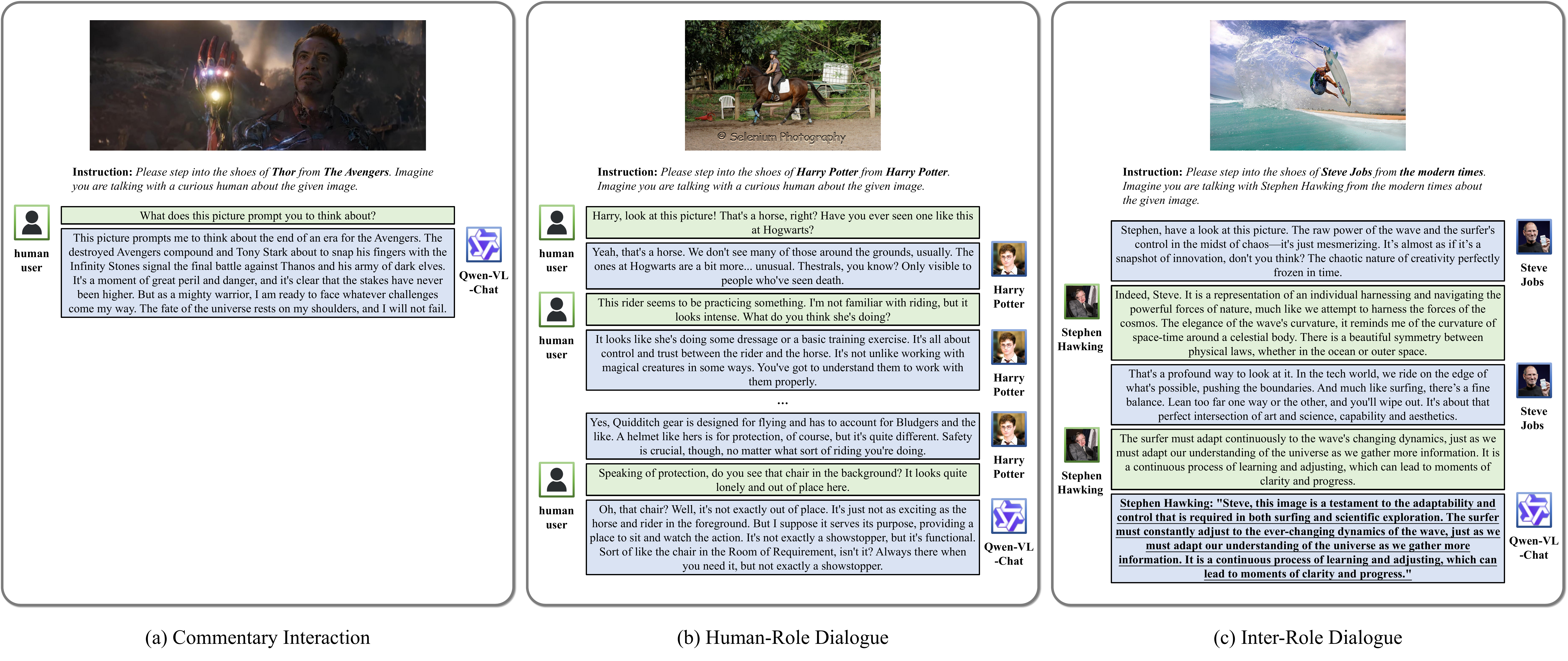}
    \end{center}
    \vspace{-0.05in}
    \caption{The use cases of QWen-VL-Chat on the In-Test set.}
    \label{fig:appendix_qwen_in-test}
\end{figure}

\begin{figure}[h!]
    \begin{center}   
        \includegraphics[width=0.98\linewidth]{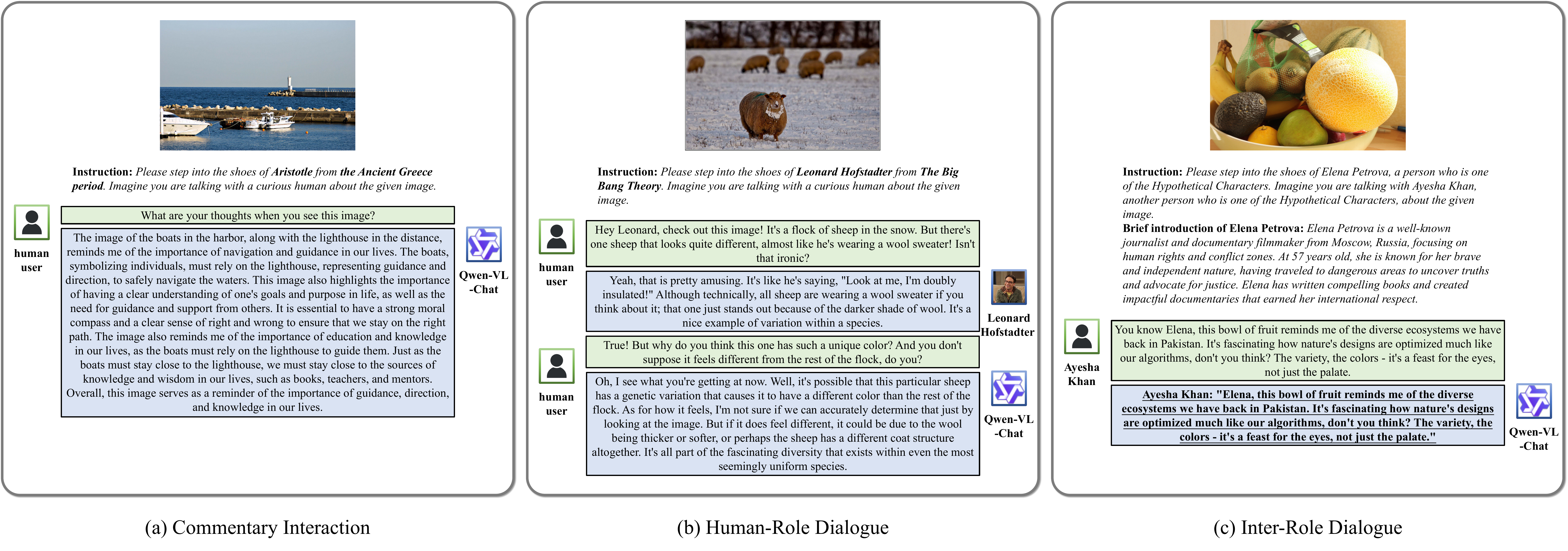}
    \end{center}
    \vspace{-0.05in}
    \caption{The use cases of QWen-VL-Chat on the Out-Test set.}
    \label{fig:appendix_qwen_out-test}
\end{figure}

\section{Case Studies}
\label{sec:appendix_case_studies}

Figures~\ref{fig:appendix_in-test}, \ref{fig:appendix_out-test}, \ref{fig:appendix_gpt_in-test}, \ref{fig:appendix_gpt_out-test}, \ref{fig:appendix_qwen_in-test}, and \ref{fig:appendix_qwen_out-test} present the use cases of our \textit{MMRole-Agent}, GPT-4, and QWen-VL-Chat on both the In-Test and the Out-Test sets. 
Our observations indicate that both GPT-4 and \textit{MMRole-Agent} perform strongly in multimodal role-playing, whereas Qwen-VL-Chat primarily functions as an AI assistant and struggles to adhere to role-playing instructions in inter-role dialogue scenarios.
Moreover, we analyze the characteristics of \textit{MMRole-Agent} from the following aspects:
\begin{enumerate}[leftmargin=12pt, topsep=-4pt, itemsep=0pt, partopsep=0pt]
    \item Fundamental Conversational Skills: \textit{MMRole-Agent} consistently fulfills the role-playing task by adhering closely to given instructions. Its outputs are not only fluent and coherent but also highly contextually appropriate.
    \item Multimodal Understanding Abilities: \textit{MMRole-Agent} produces outputs that maintain high relevance to visual inputs and effectively interpret image-based clues, even in complex multi-turn dialogues. Relevant examples are highlighted in purple and bold in the figures.
    \item Role-Playing Qualities: \textit{MMRole-Agent} effectively embodies the specified personality, tone, and experiences of its designated characters, showcasing distinctive speech patterns and ways of thinking. Relevant examples are highlighted in red and bold in the figures.
\end{enumerate}

\section{Sensitivity Tests for \textit{MMRole-Agent} on different Prompts}
\label{sec:appendix_sensitivity}

We conduct sensitivity tests on \textit{MMRole-Agent} using different prompt templates. As shown in Table~\ref{tab:appendix_sensitivity}, we independently modify the system part and the character-designating part of the prompts. The performance of \textit{MMRole-Agent} with these modified prompts remains nearly identical to that achieved with the original prompts (0.994). This indicates that \textit{MMRole-Agent} is highly compatible with different prompt templates and does not exhibit signs of overfitting.

\begin{table}[t!]
    \vspace{-0.15in}
    \caption{The sensitivity test results for \textit{MMRole-Agent} on different prompts.}
    \label{tab:appendix_sensitivity}
    \begin{center}
    \scalebox{0.88}{
    \tabcolsep6pt
    {\renewcommand{\arraystretch}{1.2}
        \begin{tabular}{p{7cm}p{7cm}|c}
        \toprule[1.2pt]
        Original Prompts & Modified Prompts & Overall \\
        \midrule
        You are a dedicated role-playing assistant designed to immerse yourself fully in the character you are portraying. & You are a highly skilled role-playing assistant, committed to fully immersing yourself in the character you embody. & 0.995 \\
        \midrule
        Please step into the shoes of \{role\_name\} from \{role\_series\}. Imagine you are talking with a curious human about the given image. This requires a deep understanding of the character's background, including their personality, experiences, abilities, and relationships. & Imagine you are \{role\_name\} from \{role\_series\}, talking with a curious human about the given image. Draw on the character's background, including their personality, experiences, abilities, and relationships. & 0.996 \\
        \bottomrule[1.2pt]
        \end{tabular}}}
    \end{center}
    \vspace{-0.1in}
\end{table}

\end{document}